\def\year{2022}\relax
\documentclass[letterpaper]{article} 
\usepackage{style}
\usepackage{times}
\usepackage{helvet}
\usepackage{courier}
\usepackage[hyphens]{url}
\usepackage{graphicx}
\usepackage{natbib}
\usepackage{caption}
\DeclareCaptionStyle{ruled}{labelfont=normalfont,labelsep=colon,strut=off}
\usepackage{algorithm}
\usepackage{algorithmic}

\usepackage{newfloat}
\usepackage{listings}
\floatstyle{ruled}
\newfloat{listing}{tb}{lst}{}
\floatname{listing}{Listing}

\usepackage[utf8]{inputenc} 
\usepackage[T1]{fontenc}
\usepackage{url}            
\usepackage{booktabs}       
\usepackage{amsfonts}       
\usepackage{nicefrac}       
\usepackage{microtype}
\usepackage{xcolor}
\usepackage{graphicx}
\usepackage{float}
\usepackage{amsmath}
\usepackage{amssymb}
\usepackage{rotating}
\usepackage{tabularx}
\usepackage{balance} 
\usepackage{hyperref}
\hypersetup{
    colorlinks=true,
    citecolor=blue,
    }

\newcommand{\Eshape}{\mathbf{E}_{\boldsymbol{\alpha}}}
\newcommand{\Ealbedo}{\mathbf{E}_{\boldsymbol{\beta}}}
\newcommand{\Elight}{\mathbf{E}_{\boldsymbol{\gamma}}}
\newcommand{\Epose}{\mathbf{E}_{\boldsymbol{\theta}}}
\newcommand{\Ehair}{\mathbf{E_h}}
\newcommand{\Gshape}{\mathbf{G}_{\boldsymbol{\alpha}}}
\newcommand{\Galbedo}{\mathbf{G}_{\boldsymbol{\beta}}}
\newcommand{\Ghair}{\mathbf{G_h}}
\newcommand{\facemask}{\mathbf{M_f}}
\newcommand{\hairmask}{\mathbf{M_h}}
\newcommand{\shapecode}{\boldsymbol{\alpha}}
\newcommand{\albedocode}{\boldsymbol{\beta}}
\newcommand{\lightparam}{\boldsymbol{\hat{\gamma}}}
\newcommand{\poseparam}{\boldsymbol{\hat{\theta}}}
\newcommand{\shapeest}{\mathbf{\hat{S}}}
\newcommand{\albedoest}{\mathbf{\hat{A}}}
\newcommand{\Renderer}{\boldsymbol{\Phi}}
\newcommand{\Refiner}{\mathbf{R}}
\newcommand{\reconstface}{\mathbf{\hat{x}_f}}
\newcommand{\reconsthair}{\mathbf{\hat{x}_h}}
\newcommand{\inputimage}{\mathbf{x}}
\newcommand{\finalimage}{\mathbf{\hat{x}}}
\newcommand{\maskedface}{\mathbf{x'}}

\DeclareMathOperator*{\argmin}{arg\,min}
\newcommand{\norm}[1]{\left\lVert#1\right\rVert}
\title{MOST-GAN: 3D Morphable StyleGAN for \\Disentangled Face Image Manipulation}
\author {
    Safa C. Medin\textsuperscript{\rm 1,2} \qquad
    Bernhard Egger\textsuperscript{\rm 2,3} \qquad
    Anoop Cherian\textsuperscript{\rm 1} \qquad
    Ye Wang\textsuperscript{\rm 1} \\
    Joshua B. Tenenbaum\textsuperscript{\rm 2} \qquad
    Xiaoming Liu\textsuperscript{\rm 4} \qquad
    Tim K. Marks\textsuperscript{\rm 1}\\[6pt]
}
\affiliations {
    \textsuperscript{\rm 1}Mitsubishi Electric Research Laboratories (MERL)\\    
    \textsuperscript{\rm 2}Massachusetts Institute of Technology\\
    \textsuperscript{\rm 3}Friedrich-Alexander-University Erlangen-Nuremberg \\
    \textsuperscript{\rm 4}Michigan State University \\[6pt]
}

\begin{document}

\maketitle

\begin{abstract}
Recent advances in generative adversarial networks (GANs) have led to remarkable achievements in face image synthesis. While methods that use style-based GANs can generate strikingly photorealistic face images, it is often difficult to control the characteristics of the generated faces in a meaningful and disentangled way. Prior approaches aim to achieve such semantic control and disentanglement within the latent space of a previously trained GAN. In contrast, we propose a framework that a priori models physical attributes of the face such as 3D shape, albedo, pose, and lighting explicitly, thus providing disentanglement by design. Our method, MOST-GAN, integrates the expressive power and photorealism of style-based GANs with the physical disentanglement and flexibility of nonlinear 3D morphable models, which we couple with a state-of-the-art 2D hair manipulation network. MOST-GAN achieves photorealistic manipulation of portrait images with fully disentangled 3D control over their physical attributes, enabling extreme manipulation of lighting, facial expression, and pose variations up to full profile view.
\end{abstract}

\section{Introduction}
\label{sec:intro}
Changing certain attributes of a given portrait image, also referred to as \emph{face image manipulation}, is a popular research topic that demonstrates the synergy between computer vision and computer graphics. Face image manipulation has a wide range of applications such as varying the illumination conditions to make a portrait image more appealing~\cite{sun2019single}, changing the identity of a person to anonymize an image~\cite{gafni2019live}, and exchanging the hairstyle in a virtual try-out setting~\cite{tan2020michigan}. Two key factors make face image manipulation particularly challenging. First, the human visual system is sensitive to the smallest artifacts in synthesized face images, and careful handling of detail is therefore crucial to achieve photorealism. Second, faces are 3D objects with rich variations in shape, expression, and appearance, and inferring such 3D variations from 2D images is inherently an ill-posed problem.

StyleGAN2~\cite{Karras2019stylegan2} is currently one of the most advanced models for 2D image generation, reaching unprecedented quality and photorealism in synthesizing face images. At the same time, 3D face models, such as those based on 3D Morphable Models (3DMMs)~\cite{blanz1999morphable, egger20203d}, are commonly used in recovering 3D faces from 2D images; however, these reconstruction methods often lack photorealism~\cite{tewari2017mofa, deng2019accurate}. There are a few recent approaches that aim to combine the physically grounded modeling of 3DMMs with the synthesizing capabilities of style-based GANs~\cite{tewari2020stylerig, tewari2020pie}. However, these approaches build on a fixed generative model, StyleGAN, and apply the explicit 3D model as a guiding tool to disentangle the learned StyleGAN latent space. As a result, these models cannot escape the data manifold characterized by a trained StyleGAN. Thus, while they provide some amount of control, they lack the generalization capabilities or physical disentanglement of 3D models, which limits their ability to synthesize large variations in the physical attributes of a face image. \vspace{1mm}

\begin{figure*}[t]
    \centering
   \includegraphics[trim=0mm 0mm 0mm 0mm,clip,width=0.85\textwidth]{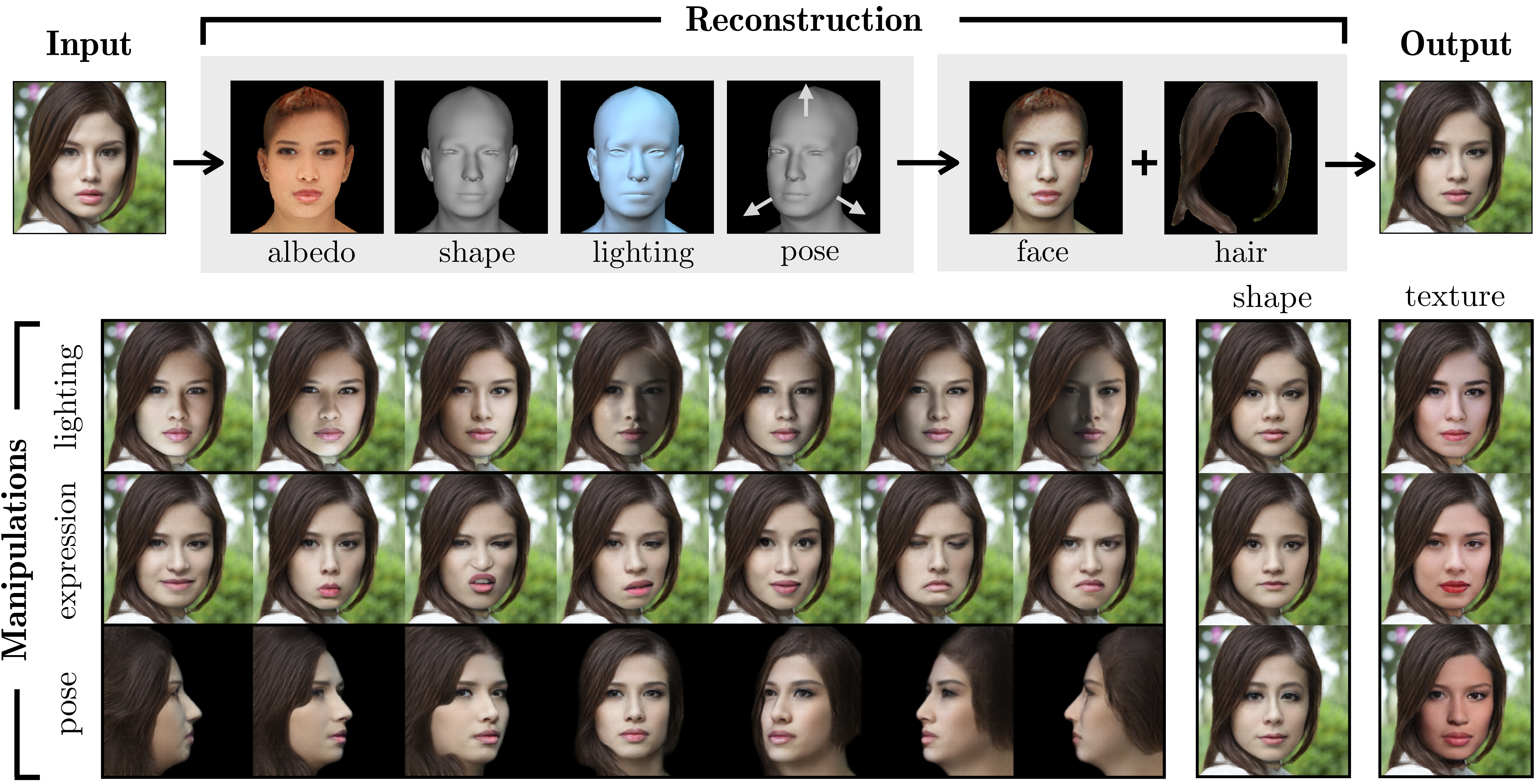} 
\caption{\small Fully disentangled, 3D controllable portrait image manipulation with MOST-GAN.}
\vspace{-2mm}
\label{fig:teaser}
\end{figure*}

In this work, we propose a nonlinear 3D face model that explicitly separates shape, albedo, lighting, and pose, which we refer to as \emph{physical attributes}. Since we represent each of these attributes explicitly, we are able to control each of them independently, either within their learned latent spaces or by direct manipulation of their 3D physical realization. By processing each physical attribute separately, our novel real-image manipulation method achieves full disentanglement of these attributes. This is in sharp contrast to state-of-the-art (SOTA) methods such as~\citet{deng2020disentangled, tewari2020stylerig, groueix2018papier}, in which entanglement among different attributes is inevitable as they are all represented in one common latent space. Our model combines the photorealism of style-based GAN architectures with the generalization capabilities of 3DMMs, which allows for extrapolating beyond the variations present in the datasets. As a result, our method is able to manipulate faces to new poses, expressions, and illumination conditions that are not well represented in the training set. We also couple our 3D face model with a SOTA 2D hair model~\cite{tan2020michigan} to achieve a complete portrait image manipulation pipeline, allowing for joint face and hair processing. The contributions of this work include:
\begin{itemize}
    \item We present a novel face image manipulation method, {3D \textbf{MO}rphable \textbf{ST}yle\textbf{GAN}} (MOST-GAN), which by design achieves full disentanglement of shape, albedo, lighting, pose, and hair.
    \item We successfully combine the generalization capabilities of 3DMMs with the photorealism of style-based GANs, which enables us to synthesize novel 3D-grounded portrait images with extreme variations that are rare or nonexistent in the training data. 
    \item We develop a 3D-guided 2D hair manipulation algorithm, allowing for photorealistic and consistent hair styles and appearances over pose variations up to full profile views.
    \end{itemize}

\section{Related Work}
\label{sec:related-work}
\textbf{Generative adversarial networks.} Generative adversarial networks (GANs)~\cite{goodfellow2014generative} have set new standards in photorealistic image generation, with recent style-based methods StyleGAN~\cite{karras2019style} and StyleGAN2~\cite{Karras2019stylegan2} generating faces that are barely distinguishable from real photos. 
As conventional GANs learn only 2D representations, several works propose 3D GANs to achieve better understanding of the 3D world, via voxel-based~\cite{choy20163d,wu2016learning,wu2017marrnet,zhu2018visual,nguyen2019hologan,xie2019pix2vox,nguyen2020blockgan,lunz2020inverse} or mesh-based representations~\cite{wang2018pixel2mesh, groueix2018papier, pan2019deep}.  
Recently, neural implicit representations have facilitated continuous 3D scene synthesis, including 3D faces~\cite{schwarz2020graf, chanmonteiro2020pi-GAN}. These methods, however, allow only limited control of facial pose. In another line of work, the 3D scene information is extracted from 2D GANs such as StyleGAN2 to manipulate 2D images in 3D~\cite{shen2020closed, harkonen2020ganspace} and recover explicit 3D shapes from images~\cite{pan20202d, zhang2020image}. However, these methods do not employ strong shape priors such as 3DMMs, limiting their 3D manipulation capabilities. In contrast, we start from a 3D architecture while incorporating StyleGAN2 inside our network, which we train without using real 3D data.

\textbf{3D Morphable Models.} There is a classic line of research based on 3D Morphable Models (3DMMs)~\cite{blanz1999morphable,egger20203d} that aims for an object-specific 3D model for faces based on high-quality 3D scans. Conventional linear 3DMMs such as the Basel Face Model~\cite{paysan20093d, gerig2018morphable} and FLAME~\cite{FLAME:SiggraphAsia2017} typically suffer from a lack of expressiveness, due to their simplistic PCA-based texture and shape models and limited training data. 
To improve the representational power of 3DMMs, \citet{tran2018nonlinear, tran2019learning,tran2019towards} proposed a nonlinear 3DMM that achieves better reconstruction quality than linear 3DMMs. 
Nonlinear models based on deep neural networks have also been used for realistic texture synthesis for various tasks~\cite{saito2017photorealistic, slossberg2018high, nagano2018pagan}. 
In this work, we build our face model as a nonlinear 3DMM based on the FLAME topology. Although the linear bases of FLAME do not yield photorealistic images, we use them to generate synthetic images for pretraining and to regularize our albedo reconstructions. In contrast to~\citet{tran2018nonlinear}, we use separate encoders for different face attributes to foster further disentanglement among them, and employ StyleGAN2 for albedo synthesis, which generates images with better photorealism.

\begin{figure*}[t]
\centering
    \includegraphics[trim=0mm 0mm 0mm 0mm,clip,width=0.85\textwidth]{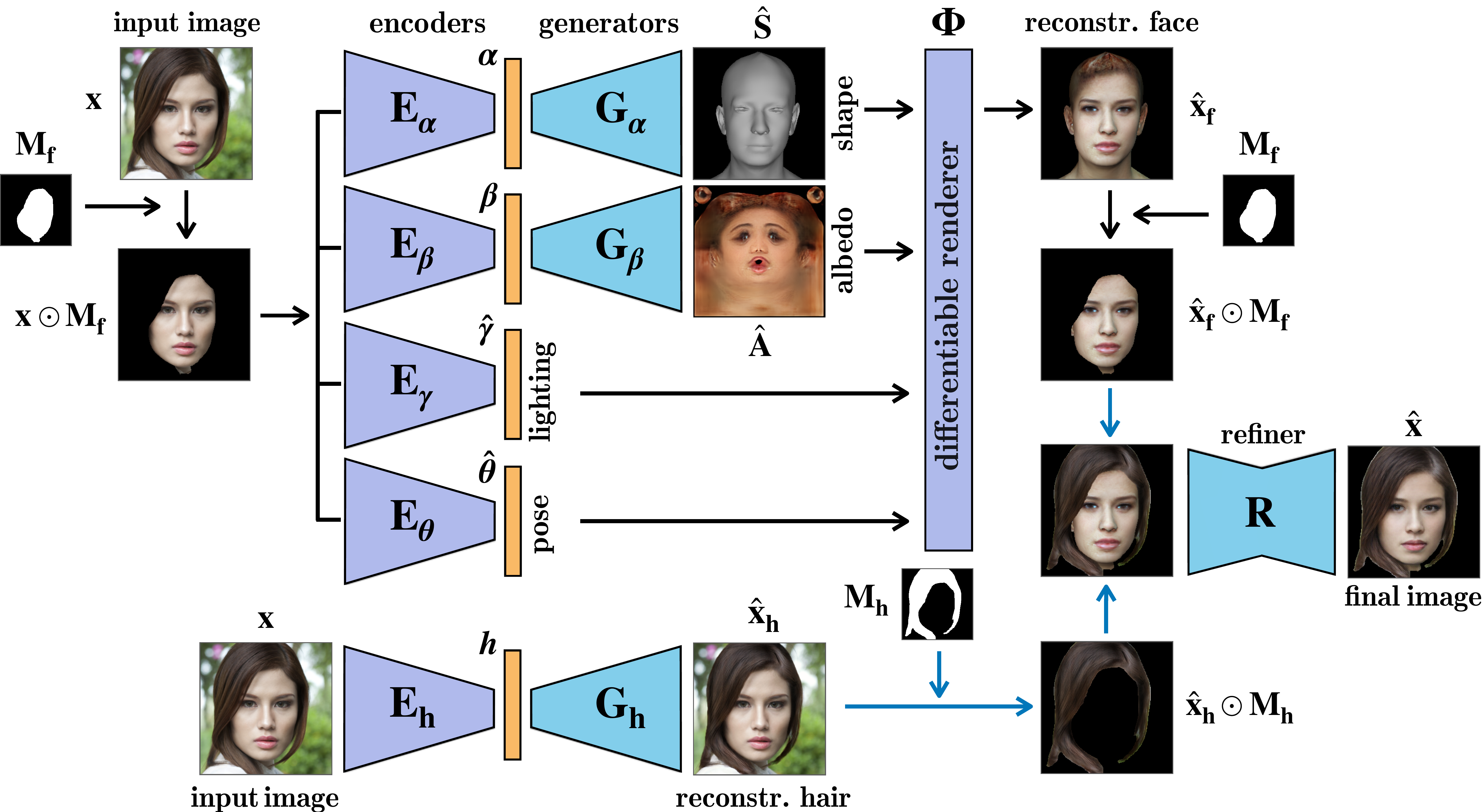}
\caption{\small \textbf{Overview of our architecture.} Our model starts with a set of encoders for shape, albedo, lighting, pose, and hairstyle given an input image. To reconstruct the shape and albedo in their physical spaces, we use a convolutional generator for shape and a StyleGAN2 architecture for albedo. A hair generator reconstructs the hair in 2D. Reconstructed face and hair are finally fused and improved using a refiner network. All components are trained end-to-end, except for hair where we deploy a pretrained MichiGAN model \cite{tan2020michigan}.}
\vspace{-2mm}
\label{fig:overview}
\end{figure*}

\textbf{3D Face Reconstruction.}  A key application of 3DMMs is to reconstruct 3D faces from 2D images, with the objective to recover either face shape~\cite{sanyal2019learning, feng2020learning} or both shape and albedo. Methods that recover both shape and albedo have benefited from advancements in GANs, which enable higher quality and more realistic texture synthesis~\cite{slossberg2018high, gecer2019ganfit, lattas2020avatarme}. Among these approaches, GANFIT~\cite{gecer2019ganfit} and AvatarMe~\cite{lattas2020avatarme} obtain face reconstructions with high-frequency details, but they require large 3D datasets for training. 
Unlike those methods, ours does not rely on high-quality 3D data for photorealism---instead, we learn to generate detailed 3D face representations from 2D face images. Several other methods also recover 3D faces from only 2D images~\cite{tewari2017mofa, deng2019accurate}, although their reconstructions cannot be used for manipulating faces due to the lack of photorealism and missing details such as hair or teeth.

\textbf{Face image manipulation.} Recent research has aimed to combine 3DMMs with SOTA GANs to edit portrait images in a disentangled manner~\cite{usman2019puppetgan, kowalski2020config, deng2020disentangled, tewari2020stylerig, tewari2020pie, ghosh2020gif, buhler2021varitex, piao2021inverting}. Among them, DiscoFaceGAN~\cite{deng2020disentangled} promotes disentanglement between face attributes via contrastive learning, while StyleRig~\cite{tewari2020stylerig} couples a 3DMM with a pretrained 2D StyleGAN and manipulates images in the latent space of the StyleGAN. Since both methods rely on 2D generative networks, they are not able to handle extreme variations in 3D such as extreme lighting, facial expression, or pose. Furthermore, to manipulate real images, they must be embedded into the learned latent spaces via Image2StyleGAN~\cite{abdal2019image2stylegan}, which hinders the quality of results. 
To circumvent such issues, Portrait Image Embedding (PIE)~\cite{tewari2020pie} introduces a novel optimization algorithm to embed real images into the latent space while preserving their photorealism. However, since both StyleRig and PIE are built on a pretrained 2D StyleGAN, the learned latent space limits them to variations that are well represented in the FFHQ dataset~\cite{karras2019style}. 
Further, since these methods aim to disentangle their latent spaces {\it post hoc}, full disentanglement between physical attributes cannot be attained. 
In work concurrent to our research, GAR~\cite{piao2021inverting} proposes a realistic face reconstruction method that is used to manipulate portrait images, and VariTex~\cite{buhler2021varitex} introduces a variational texture generator to synthesize realistic face images while achieving control over them. 
Both of these methods, however, provide only head pose and expression manipulation results, and similar to the other methods presented in this paragraph, they do not show results with large pose variations (larger than $45^\circ$).

\textbf{Societal impacts.} We envision that our method could be used in numerous applications including creative uses in the entertainment sector, generation of realistic training data, or anonymization of public images. We have seen previous related methods misused to produce malicious content, such as fake news, and our method could enable face editing with larger variations. However, the research community is simultaneously creating methods to detect and mitigate such applications~\cite{ciftci2020fakecatcher}, and legal regulations to prohibit such misuse are under consideration.

\section{Methods}
Our approach combines a statistical model of 3D faces with a style-based GAN, achieving a realistic and fully disentangled 3D model of faces. We achieve such disentanglement by individually processing each of the face's physical attributes and hair in the architecture, through separate encoders and decoders, as shown in Fig.~\ref{fig:overview}. Such explicit control enables us to extrapolate beyond what is well represented in the training set, allowing for face synthesis in extreme poses, facial expressions, and lighting conditions.

\subsection{Problem Formulation}

Our face image manipulation method relies on reconstructing accurate and photorealistic 3D faces from 2D images using the architecture shown in Fig.~\ref{fig:overview}. 
Here, we assume that a portrait image can be decomposed into five different attributes: four \emph{physical attributes} (3D shape, albedo, lighting, and pose), and hair. 
Our face model employs a set of encoders $\{ \Eshape, \Ealbedo, \Elight, \Epose \}$. 
Given a masked face image \smash{$\maskedface := \inputimage \odot \facemask$}, where $\inputimage$ denotes the input image and $\facemask$ denotes its estimated face mask~\cite{chen2017rethinking}, the encoders $\Eshape$ and $\Ealbedo$ extract a latent shape code $\shapecode$ and albedo code~$\albedocode$, while $\Elight$ and $\Epose$ directly estimate the lighting parameters \smash{$\lightparam$} and pose parameters \smash{$\poseparam$}. To generate a face image, the shape and albedo codes are fed to a shape generator $\Gshape$ and albedo generator $\Galbedo$, respectively, to produce a 3D shape~\smash{$\shapeest$} and albedo map \smash{$\albedoest$}. Next, a differentiable renderer~$\Renderer$ renders the generated 3D model \smash{$\{ \shapeest, \albedoest \}$} using the lighting and pose parameters \smash{$\{\lightparam, \poseparam\}$} to produce the reconstructed face~$\reconstface$: \smash{$\reconstface = \Renderer(\shapeest, \albedoest, \lightparam, \poseparam)$}. 

Our hair model consists of an encoder $\Ehair$ and a generator $\Ghair$ to produce a portrait image with reconstructed hair $\reconsthair$. Finally, the outputs of the face model and the hair model are combined using a face mask $\facemask$ and a hair mask $\hairmask$, then passed through a refiner network $\Refiner$ that produces the final image $\finalimage$. Formally,
given a set of $N$ portrait images along with their face masks and hair masks \smash{$\{(\inputimage^i, \mathbf{M}_{\mathbf{f}}^i, \mathbf{M}_{\mathbf{h}}^i)\}_{i=1}^N$}, our objective is to solve the following optimization problem:
\begin{equation}
    \argmin_{ \{ \Eshape, \Ealbedo, \Elight, \Epose, \Gshape, \Galbedo, \Refiner  \}} \sum_{i=1}^N \norm {\inputimage^i \odot (\mathbf{M}_{\mathbf{f}}^i + \mathbf{M}_{\mathbf{h}}^i) - \finalimage^i}_1
\end{equation}
where each final image $\finalimage = \Refiner(\reconstface \odot \facemask + \reconsthair \odot \hairmask$), with $\reconstface = \Renderer (\Gshape(\Eshape(\maskedface)), \Galbedo(\Ealbedo(\maskedface)), \Elight(\maskedface), \Epose(\maskedface) )$ and $\reconsthair = \Ghair(\Ehair(\inputimage))$. In later sections, we will show that adopting this objective enables us to edit portrait images in a fully disentangled manner while preserving their photorealism.

\subsection{Face Model}

Our face model, demarcated in Fig.~\ref{fig:overview} by black connecting arrows, consists of four physical attribute encoders, two generators, and a differentiable renderer~\cite{ravi2020pytorch3d}. In the shape pipeline, the shape code $\shapecode$ is input to a convolutional generator, $\Gshape$. The generated 3D shape, \smash{$\shapeest$}, is composed of 3 channels in the UV-space that represent the 3D coordinates of vertices~\cite{tran2018nonlinear} by their displacement from the FLAME mean head model. In parallel, the albedo code~$\albedocode$ goes through a StyleGAN2~\cite{Karras2019stylegan2} generator~$\Galbedo$ that outputs an RGB albedo map \smash{$\albedoest$} in the UV-space. Since most of the variations in face images are due to the variations in the albedo, generating albedo with a style-based architecture is a crucial step to achieve realism in the final output. Furthermore, in order to allow for more expressive latent spaces of shape and albedo, we let our model learn them without being constrained to the subspace defined by the original 3DMM. Finally, we represent the estimated lighting \smash{$\lightparam$} using a spherical harmonics parameterization with 3 bands~\cite{ramamoorthi2001efficient, zhang2006face}, and our 6-DOF pose vector~\smash{$\poseparam$} includes $3$ parameters for 3D rotation using the axis-angle representation and $3$ parameters for 3D translation.

We divide our training process into two stages: 1) we pretrain our face model on synthetically generated faces; then 2) we generalize our model to real faces by training on real 2D images. The loss functions for each stage are introduced in the equations below and the subsequent explanations:\\

\noindent
\begin{minipage}[c]{\columnwidth}
\centering
\vspace{-4pt}
\textbf{Synthetic data Pretraining}
\begin{align}
    &L_{\mathrm{image}}^{\mathrm{syn}} \hspace{-24pt} &&= \Vert \mathbf{x} - \reconstface \Vert_2^2 \label{eq:image-synth} \\
    &L_{\mathrm{albedo}}^{\mathrm{syn}} \hspace{-24pt} &&= \Vert \mathbf{A} - \mathbf{\hat{A}} \Vert_2^2 \label{eq:alb-synth} \\
    &L_{\mathrm{shape}}^{\mathrm{syn}} \hspace{-24pt} &&= \Vert \mathbf{w_s}^T(\mathbf{S} - \mathbf{\hat{S})\,} \Vert_2^2 \label{eq:shape-synth} \\
    &L_{\mathrm{pose}}^{\mathrm{syn}} \hspace{-24pt} &&= \Vert \boldsymbol{\theta} - \boldsymbol{\hat{\theta}}  \Vert_2^2 \label{eq:pose-synth} \\
    &L_{\mathrm{lighting}}^{\mathrm{syn}} \hspace{-24pt} &&= \Vert \boldsymbol{\gamma}  - \boldsymbol{\hat{\gamma}}  \Vert_2^2 \label{eq:gamma-synth}\\
    &L_{\mathrm{reg}}^{\mathrm{syn}} \hspace{-28pt} &&= \lambda_\alpha \Vert \shapecode \Vert_2^2 + \lambda_\beta \Vert \albedocode \Vert_2^2 \qquad \label{eq:reg-synth}
\end{align}
\end{minipage}

\noindent
\begin{minipage}[c]{\columnwidth}
\centering
\vspace{6pt}
\textbf{Real data Training}
\begin{align}
    &L_{\mathrm{image}}^{\mathrm{real}} \hspace{-10pt} &&= \Vert  \mathbf{x} \odot \mathbf{M_f} - \mathbf{\hat{x}} \odot \mathbf{M_f} \Vert_2^2 \label{eq:image-real} \\
    &L_{\mathrm{identity}}^{\mathrm{real}} \hspace{-10pt} &&= 1 - \cos( f_\mathrm{id}(\mathbf{x}),\, f_\mathrm{id}(\mathbf{\hat{x}')}) \label{eq:id-real}\\
    &L_{\mathrm{landmark}}^{\mathrm{real}} \hspace{-10pt} &&= \Vert\mathbf{w_l}^T[ f_\mathrm{lmk}^{(1)} (\mathbf{x}) - f_\mathrm{lmk}^{(2)} (\mathbf{\hat{S}})]\, \Vert_2^2 \label{eq:lmk-real} \\
    &L_{\mathrm{albedo}}^{\mathrm{real}} \hspace{-10pt} &&= \Vert (\mathbf{B^TB})^{-1} \mathbf{B^T} ( \mathbf{\hat{A}} - \mathbf{\bar{A}}) \Vert_2^2 \label{eq:alb-real} \\ 
    &L_{\mathrm{lighting}}^{\mathrm{real}} \hspace{-10pt} &&= (\boldsymbol{\hat{\gamma}} - \boldsymbol{\bar{\gamma}})^T \mathbf{\Sigma}^{-1} (\boldsymbol{\hat{\gamma}} - \boldsymbol{\bar{\gamma}}) \label{eq:gamma-real}\\
    &L_{\mathrm{reg}}^{\mathrm{real}} \hspace{-18pt} &&= \lambda_\alpha \Vert \shapecode \Vert_2^2 + \lambda_\beta \Vert \albedocode \Vert_2^2 \label{eq:reg-real}
\end{align}
\end{minipage} \\

\textbf{Pretraining on Synthetic Data.} The first stage is a pretraining step to allow our network to capture important characteristics of faces using strong supervision coming from a linear 3DMM. 
In this stage, we use the FLAME model to sample $80,000$ faces under an illumination and pose prior~\cite{deng2020disentangled}. We translate each face in 3D so that the rendered faces have the same 2D alignment as the FFHQ faces. Although these synthetic faces lack realism, they have ground truth values for the disentangled physical attributes albedo $\mathbf{A}$, shape $\mathbf{S}$, pose $\boldsymbol{\theta}$, and lighting $\boldsymbol{\gamma}$, which we use to guide pretraining. Our loss function for pretraining consists of three parts: reconstruction losses for the reconstructed face image~\eqref{eq:image-synth} and for the four physical attributes~\eqref{eq:alb-synth}--\eqref{eq:gamma-synth}; regularization for shape and albedo codes~\eqref{eq:reg-synth}; and a non-saturating logistic GAN loss~\cite{goodfellow2016nips} to improve photorealism. In the shape reconstruction loss~\eqref{eq:shape-synth}, we introduce a weighting term $\mathbf{w_s}$ to upweight vertices in regions surrounding salient facial features (e.g., eyes, eyebrows, mouth).

\textbf{Training on Real Data.} After pretraining, we train our model using the FFHQ face dataset~\cite{karras2019style}, where for simplicity we eliminate the images with glasses. We obtain the face mask~$\mathbf{M_f}$ for each image automatically using a semantic segmentation network~\cite{or2020lifespan, chen2017rethinking}, then feed the masked 2D face images to the network. We train our face model in an end-to-end fashion, where we combine the loss functions in~\eqref{eq:image-real}--\eqref{eq:reg-real} with a non-saturating logistic GAN loss. Since we do not know the ground truth physical attributes for the real face images, we cannot apply any of the physical attribute reconstruction losses \eqref{eq:alb-synth}--\eqref{eq:gamma-synth}. The only reconstruction loss we apply is a pixelwise reconstruction loss for the masked faces~\eqref{eq:image-real}. Defining the full reconstructed image as $\mathbf{\hat{x}}' :=  \mathbf{x} \odot (1-\mathbf{M_f})  + \mathbf{\hat{x}} \odot \mathbf{M_f}$, we impose an identity loss~\eqref{eq:id-real}, where $f_\mathrm{id}(\cdot)$ denotes the feature vector extracted by the Arcface face recognition network~\cite{deng2018arcface}, and $\cos(\cdot, \cdot)$ denotes cosine similarity. Our landmark loss~\eqref{eq:lmk-real} measures the distance between the image-plane projections of the 3D facial landmark locations in the input image (estimated using~\cite{bulat2017far}) and the corresponding locations in the reconstructed 3D shape model. The shape model vertices corresponding to specific facial landmarks are defined by the FLAME topology, and the weighting term $\mathbf{w_l}$ places more weight on important landmarks such as the lip outlines to keep our learned model faithful to the FLAME topology.

Since the decomposition of an input image into physical face properties is an ill-posed problem, there are ambiguities such as the relative contributions of color lighting intensities and surface albedo to the RGB appearance of a skin pixel. To help resolve this ambiguity, we introduce an albedo regularization loss~\eqref{eq:alb-real} to minimize the projection of our reconstructed albedo into the FLAME model's albedo PCA space.
Here, $\mathbf{\bar{A}}$ and $\mathbf{B}$ respectively represent the mean and basis vectors of the FLAME albedo model. To address the same ambiguity, we also include a lighting regularization loss~\eqref{eq:gamma-real}, which minimizes the log-likelihood of the reconstructed lighting parameters $\boldsymbol{\hat{\gamma}}$ under a multivariate Gaussian distribution over lighting conditions. To obtain that distribution, we sampled 50,000 lighting vectors using the prior provided by~\citet{deng2020disentangled} and calculate their sample mean $\boldsymbol{\bar{\gamma}}$ and sample covariance $\mathbf{\Sigma}$. As in pretraining, \eqref{eq:reg-real} regularizes the shape and albedo codes.

\begin{figure}[t]
\begin{center}
    \includegraphics[trim=0mm 0mm 0mm 0mm,clip,width=1.0\columnwidth]{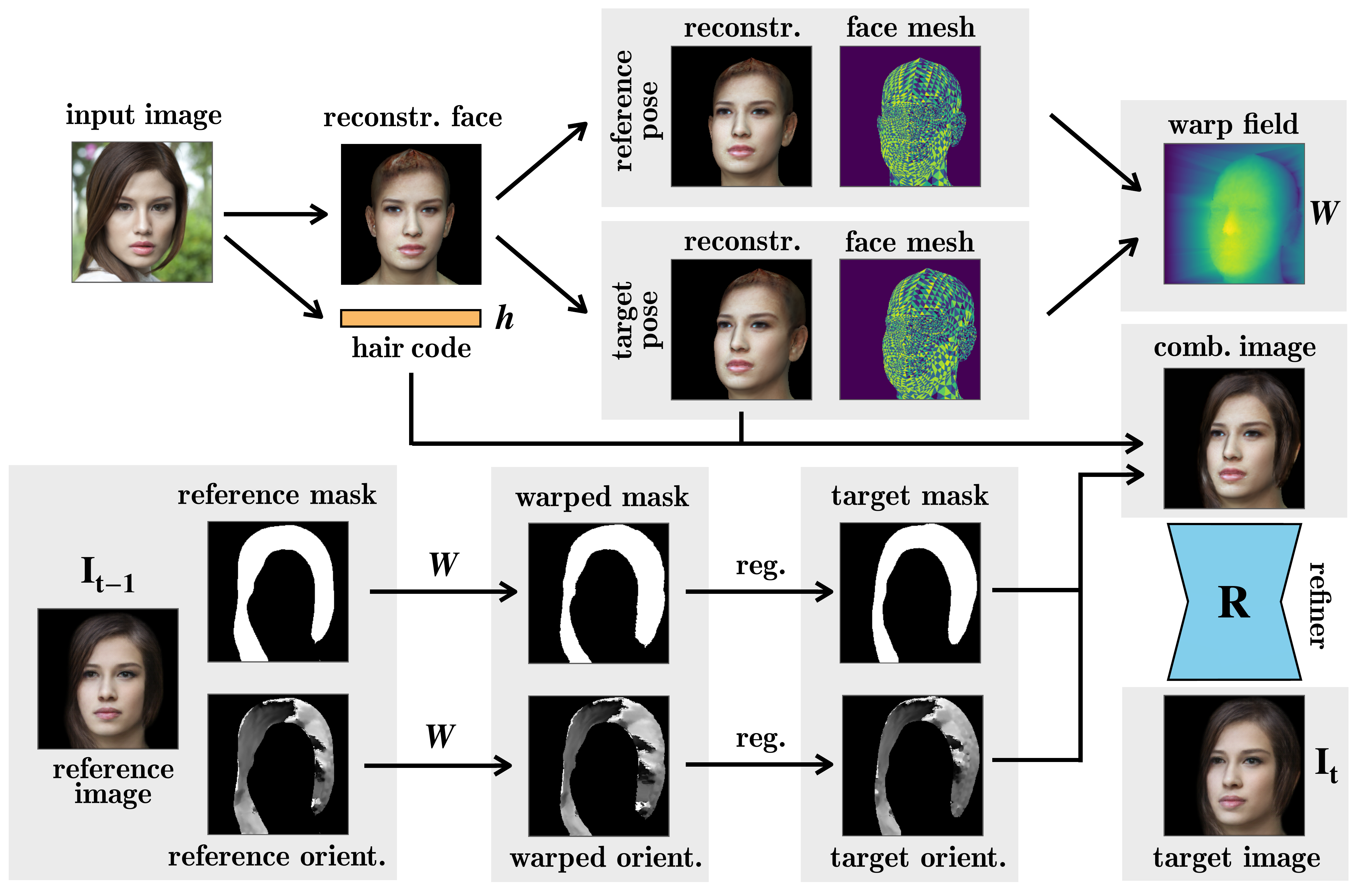}
\end{center}
\caption{\small \textbf{One iteration of our hair manipulation algorithm}. Given a reference pose from the previous iteration and a target pose, we calculate a 2D warp field based on how 3D vertices move within the image plane. Given a reference image from the previous iteration $\mathbf{I_{t-1}}$ along with its reference mask and orientation, we use this warp field to warp the mask and the orientation, which we regularize to obtain the target mask and orientation. Next, we combine these with the hair appearance code obtained from the original input image and the reconstructed face reposed to the target pose, to obtain a novel portrait image $\mathbf{I_{t}}$. At the end, we feed this image through the refiner to obtain a photorealistic output. This algorithm is invoked sequentially starting from the original pose. The elements shown on gray backgrounds are updated in each iteration.}
\vspace{-5mm}
\label{fig:hair-warp}
\end{figure}

\subsection{Hair Model}
\label{sec:hair-model}
Since hair has a more complex structure than faces, representing and manipulating hair in 3D is a very challenging problem. This motivates us to manipulate hair in 2D, but to couple the hair generation process with our 3D face model. We build our hair model upon a state-of-the-art 2D model, MichiGAN~\cite{tan2020michigan}, which disentangles hair shape, structure, and appearance by processing them separately and combines them with a backbone network. Here, shape refers to a 2D binary mask of the hair region, structure is represented as a 2D hair strand orientation map, and appearance refers to the global color and style of the hair which is encoded in a latent space. We incorporate a pretrained MichiGAN in our training pipeline, which we briefly represent as an encoder-decoder style model in Fig.~\ref{fig:overview}. When we repose faces at inference time, we couple MichiGAN with our 3D face model to change the shape and structure of the hair without changing its appearance code. \vspace{3mm}

\textbf{Coupling with Face Model.} Our 3D-guided hair manipulation algorithm is illustrated in Fig.~\ref{fig:hair-warp}. Since our face model reconstructs explicit 3D face shapes, we use these to reason about how the hair will move in 2D by calculating a 2D warp field{~\cite{li2019dense}}. We derive the 2D warp field based on the pose-induced movement of the 3D face vertices, then extrapolate the face's warp field to the rest of the image.

We use the warp field to warp the hair mask and the hair orientation map in 2D. Since this process can introduce warp artifacts, however, we regularize the warped masks by projecting them onto a PCA basis calculated from a dataset of binary hair masks of portrait images. In addition to obtaining hair masks from the FFHQ dataset~\cite{karras2019style}, we extract hair masks from the USC HairSalon database~\cite{hu2015single} by rendering that dataset's 3D hair models with faces in extreme poses to allow for accurate and consistent hair masks under large pose variations. The orientation map, on the other hand, is regularized as part of the MichiGAN framework, which outputs a map that is consistent with the warped map and aligned with the regularized hair mask. Finally, the reconstructed face in the target pose, hair appearance code, hair mask, and hair orientation map are combined by the MichiGAN pipeline to produce the reposed portrait image, which is then processed with the refiner (described below). For large pose variations, we invoke this algorithm sequentially by going from reference pose to target pose in multiple steps, and we regularize the warped masks and orientation maps at each step. For more details, please see the supplementary material.

\subsection{Refinement}
Although our combined model's rendered 3D face reconstructions and 2D hair reconstructions closely resemble the original images, there is still a small realism gap that needs to be filled. In particular, since we regularize the reconstructed albedos using the FLAME albedo space, the reconstructions do not exhibit sufficient variation in the eye regions, {and they lack certain details such as eyelashes, facial hair, teeth, and accessories, which are not modeled by the FLAME mesh template.} Furthermore, since face and hair are processed separately, some reconstructions have blending issues between the face and the hair. To address these issues, we utilize a refiner network, which closes the realism gap between the reconstructions and the original images while making only a minimal change to the reconstructions. We employ a U-Net~\cite{ronneberger2015u} that takes in an image combining the reconstructed face and hair and outputs a more realistic portrait image, as shown in Fig.~\ref{fig:overview}.

After freezing the weights of the rest of the model, we train the refiner with pairs of original images from the dataset and reconstructed images. For the refiner, we combine the same adversarial loss and identity loss~\eqref{eq:id-real} described above with a reconstruction loss based on the VGG-16 perceptual loss ~\cite{simonyan2014very, johnson2016perceptual}, promoting better reconstruction quality for hair.

\begin{figure}[t]
    \centering
    \includegraphics[trim=0mm 0mm 0mm 0mm,clip,width=1.0\columnwidth]{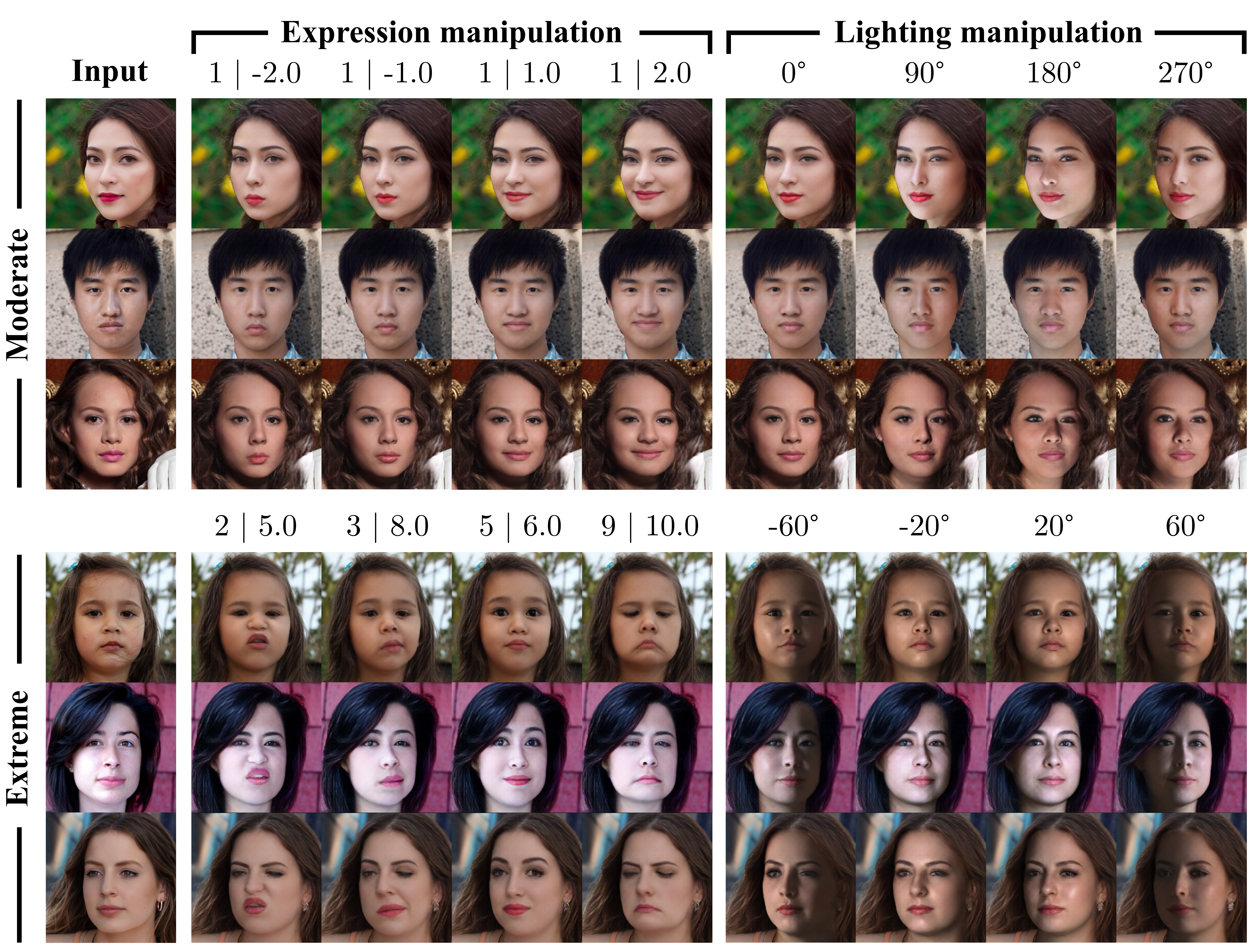} 
    \caption{\small Expression and lighting manipulation results. \textbf{Expression manipulation (\emph{left}).} We illustrate both moderate (\emph{top}) and extreme (\emph{bottom}) expression variations. The two numbers above each column indicate which FLAME expression eigenvector is used and by how many standard deviations it is scaled. \textbf{Lighting manipulation (\emph{right}).} \emph{Top:} For moderate variation, we rotate the reconstructed lighting around the camera axis by the angle above each column. \emph{Bottom:} For extreme lighting variation, we render the reconstructed 3D model using a point light source and Phong shading model.}
    \vspace{-4mm}
    \label{fig:expr-light-manip}
\end{figure}

\begin{figure}[t]
    \centering
    \includegraphics[width=0.48\textwidth]{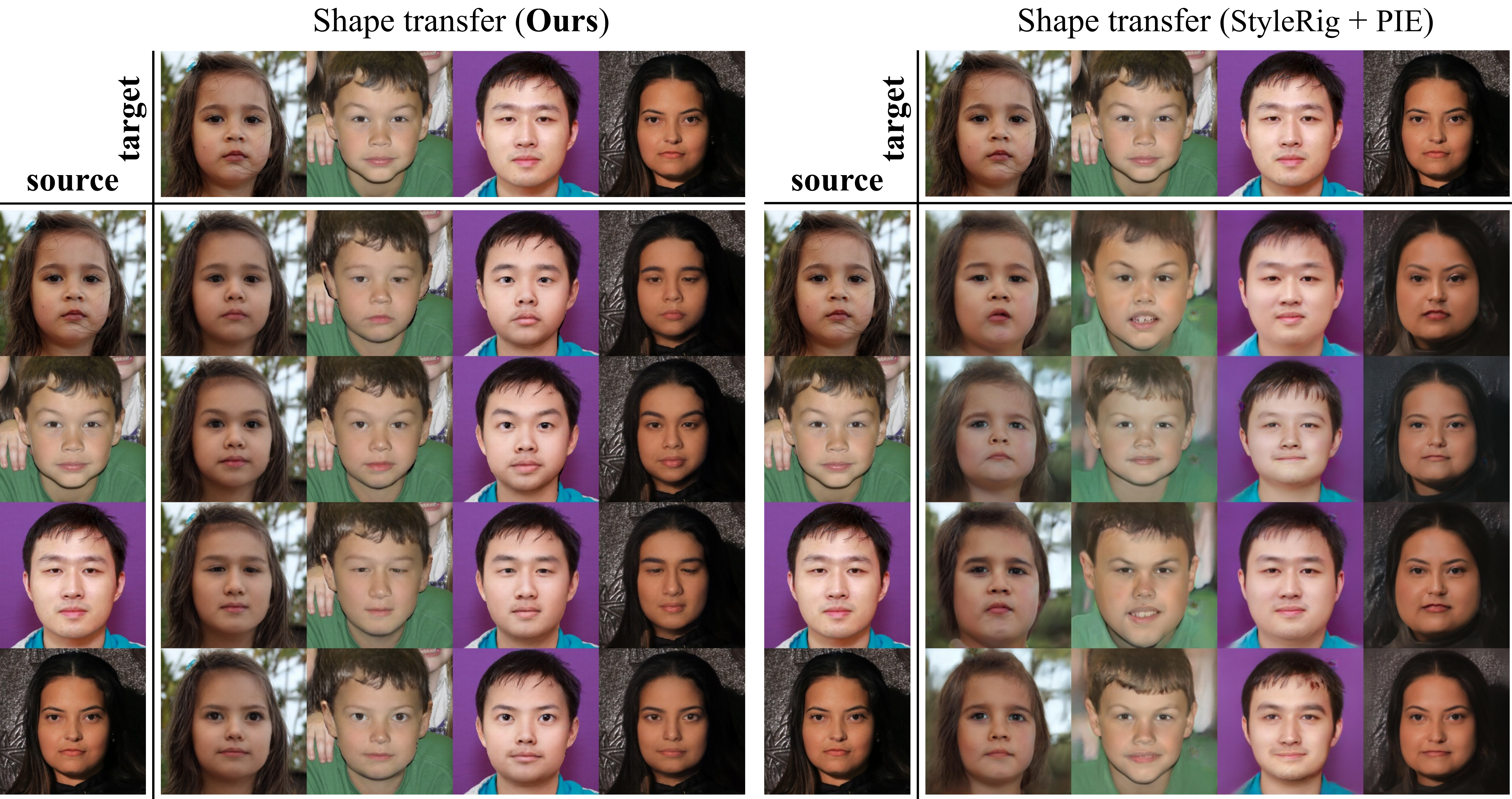}
    \caption{\small Shape transfer comparison. We transfer the 3D shape of each source image to each target image while keeping everything else unchanged. Our results (\emph{left}) demonstrate more accurate shape transfer and much better disentanglement between shape and other attributes (e.g., albedo, pose, and hair) {than the combination of StyleRig~\cite{tewari2020stylerig} and PIE~\cite{tewari2020pie}} (\emph{right}). }
    \vspace{-4mm}
    \label{fig:shape-transfer}
\end{figure}

\section{Experiments and Results}
\label{sec:experiments}

In our experiments, we manipulate portrait images with respect to several physical attributes and compare them with a SOTA real-image manipulation method, PIE~\cite{tewari2020pie}. Besides providing qualitative comparisons of the two methods, we also quantitatively compare the performance of our pose editing algorithm by employing a head pose estimator~\cite{ruiz2018fine} to measure the error between the desired and estimated head poses. 

Because MOST-GAN generates a full 3D model, we can manipulate physical attributes beyond the distribution of the training set. We can also modify the face in ways not anticipated during training, such as relighting faces using a different lighting and shading model.

\textbf{Expression and lighting manipulation.} We illustrate our facial expression and lighting manipulation results in Fig.~\ref{fig:expr-light-manip}. To edit facial expression (left), we choose an eigenvector from the FLAME expression basis and multiply it by a constant factor to obtain an offset, which we add to the vertex locations in our model's reconstructed 3D shape. In the moderate examples (top left), we use the first eigenvector to add smile/frown variations up to $\pm 2$ standard deviations. In the extreme examples (bottom left), we scale $5$ different expression eigenvectors by up to $10$ standard deviations. For lighting manipulation (right), the moderate edits (top right) rotate the reconstructed lighting around the camera axis (axis perpendicular to the image plane). For extreme lighting variations (bottom right), we employ a point light source and Phong shading model, where we rotate the light source horizontally around the vertical axis and can introduce any desired amount of specularity to the face albedo. The results demonstrate that our method easily handles extreme expressions and lighting conditions that are not well-represented in the training set and can use lighting and shading models not used in training. (We show more examples in the supplementary material.)

\begin{table*}[t!]
    \caption{\small Mean absolute errors between the desired and estimated head poses in degrees, on $100$ random images from our test set. Our method's average head pose error is significantly smaller than that of PIE~\cite{tewari2020pie}, indicating our method's superior pose disentanglement.}
\begin{tabularx}{\textwidth}{l*{13}{>{\centering\arraybackslash}X}}
\cmidrule(r){1-14}
& $\text{-}90^{\circ}$ & $\text{-}75^{\circ}$ & $\text{-}60^{\circ}$ & $\text{-}45^{\circ}$ & $\text{-}30^{\circ}$ & $\text{-}15^{\circ}$ & $0^{\circ}$ & $15^{\circ}$ & $30^{\circ}$ & $45^{\circ}$ & $60^{\circ}$ & $75^{\circ}$ & $90^{\circ}$ \\ \cmidrule(r){1-14}
\textbf{Ours} & $\mathbf{13.2}$ & $\mathbf{6.0}$ & $\mathbf{3.5}$ & $\mathbf{4.5}$ & $\mathbf{4.7}$ & $\mathbf{4.8}$ & $\mathbf{2.3}$ & $\mathbf{2.6}$ & $\mathbf{5.0}$ & $\mathbf{6.0}$ & $\mathbf{3.9}$ & $\mathbf{2.7}$ & $\mathbf{6.7}$ \\
PIE & ${\,61.7}$ & ${38.0}$ & ${22.1}$ & ${15.8}$ & ${10.8}$ & ${7.6}$  & ${3.5}$  & ${6.2}$  & ${13.1}$ & ${22.5}$ & ${35.5}$ &  ${54.8}$ &  ${81.0}$  \\  \bottomrule
\end{tabularx}
\label{tab:pose-estimation}
\end{table*}

\textbf{Shape transfer.} Our model achieves superior disentanglement of physical attributes such as shape and albedo by design, by modeling them separately and explicitly. This disentanglement is illustrated by the shape transfer results in Fig.~\ref{fig:shape-transfer}, where we transfer the 3D face shape of a source image to a target image. Our results show that our method (\emph{left}) is able to transfer the face shapes accurately, while maintaining photorealism and keeping the albedo, lighting, and hair unchanged. This is in contrast to the shape transfer results by the previous state of the art (\emph{right}, a combination of StyleRig~\cite{tewari2020stylerig} and PIE~\cite{tewari2020pie}), where for a given source shape, the transfer results have varying face shapes with noticeable differences in expressions. When the source and target images are identical (images on the diagonal), our method produces the original reconstruction by design, whereas PIE + StyleRig struggles to maintain the original identity. Our method can also transfer albedo alone, transfer multiple physical attributes (such as albedo and shape) simultaneously, and smoothly interpolate between different shapes and albedos in the latent space continuously. (See the supplementary material for examples.)

\begin{figure}[t]
    \centering
    \includegraphics[trim=0mm 0mm 0mm 0mm,clip,width=1.0\columnwidth]{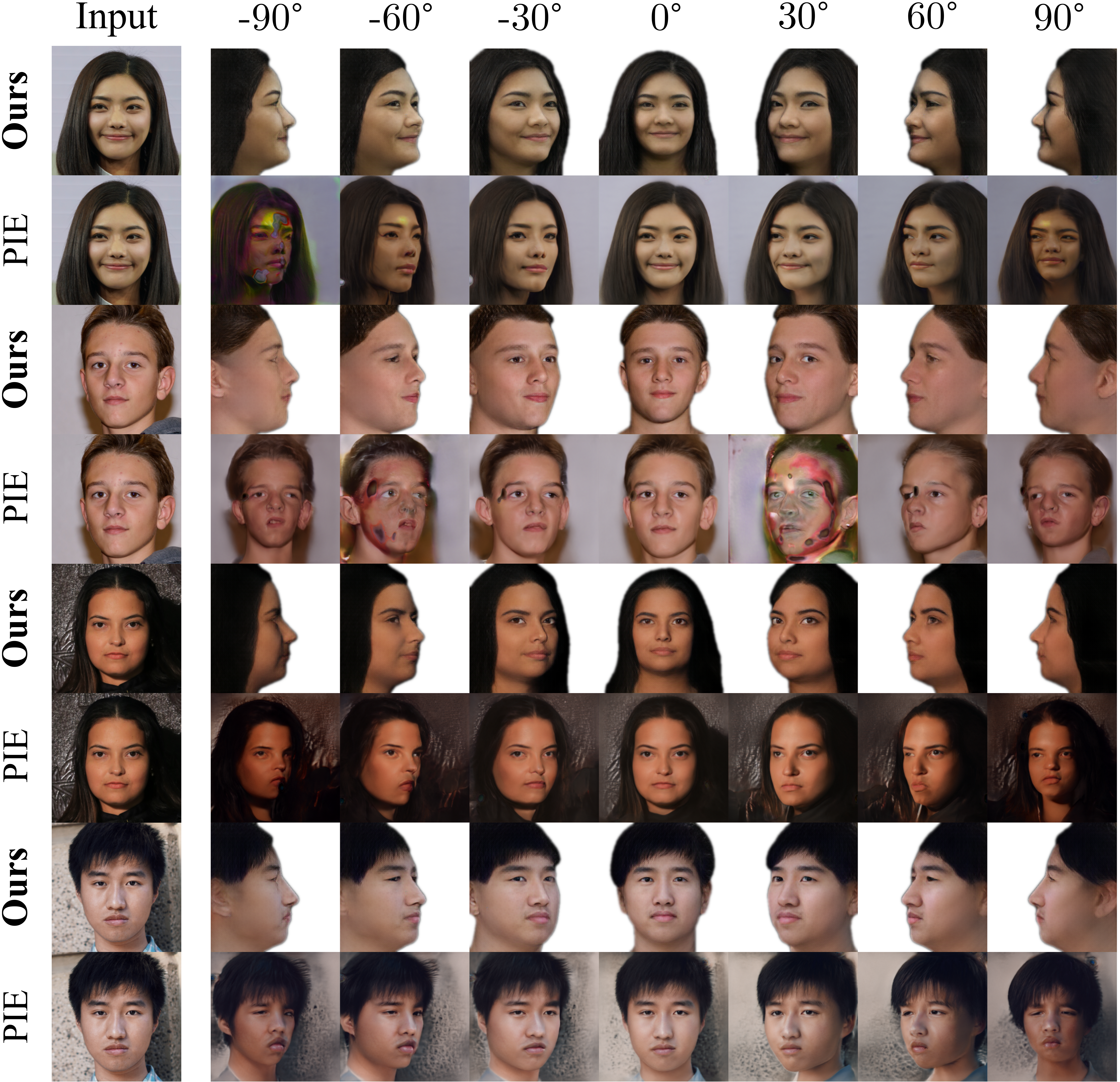}
    \caption{\small Pose manipulation results. From an input portrait image, our method accurately rotates the reconstructed 3D face all the way to profile pose. (Faces in more extreme poses appear larger due to the FFHQ alignment.) In contrast, PIE~\cite{tewari2020pie} struggles to maintain photorealism and cannot achieve large rotations.
    }
    \vspace{-4mm}
    \label{fig:pose-manip}
\end{figure}

\textbf{Pose manipulation.} In Fig.~\ref{fig:pose-manip}, we compare our pose manipulation results (odd rows) to PIE~\cite{tewari2020pie} (even rows). To edit the pose of a given portrait image, we rotate the reconstructed faces in 3D and warp the hair in 2D using our 3D-guided hair manipulation algorithm described in the Hair Model section. The results show that our method is able to rotate portrait images all the way to profile pose while keeping the identity, expression, and illumination conditions unchanged. 
For the $0^\circ$ pose, PIE~\cite{tewari2020pie} is slightly better at reconstructing the original identity. However, PIE relies on a costly optimization over the latent space of a pretrained GAN, whereas our method reconstructs 3D faces at interactive framerates ($30$ fps) using our encoder-decoder style architecture. Furthermore, PIE cannot handle extreme rotations that were not well represented in the StyleGAN training set, yielding unrealistic artifacts and an inability to achieve larger desired (target) poses. To quantify the latter, we calculate the mean absolute error between the desired and achieved head poses using a head pose estimation network~\cite{ruiz2018fine}. In particular, we randomly sampled $100$ images from our test dataset, reposed them in a range of yaw angles using our method and PIE, and calculated the average absolute pose error of each method. The results, in Table~\ref{tab:pose-estimation}, show that our method yields much more accurate pose manipulation at all pose angles.

\label{sec:limitations}
\textbf{Limitations}. Since we disentangle hair from the physical attributes by design, changing the lighting conditions has a limited effect on the hair, and that effect is achieved by the refiner. Since the hair appearance is strongly dependent on the head pose and lighting conditions, this issue could be addressed by coupling the pose and lighting with the hair model at training time. {Also}, since the reconstruction quality of hair is heavily influenced by the hair orientation map in the MichiGAN framework, achieving consistency of orientation maps over large pose variation is crucial to render photorealistic hair for reposed images. Currently, however, we handle dis-occlusions of the hair by warping the orientation maps in 2D, which sometimes yields inconsistent orientations (and thus unrealistic hair rendering) after large pose changes. {In addition, our model tends to attribute skin color mostly to the lighting component, which is due to the fact that samples from the FLAME albedo basis, which we use to regularize our albedo reconstructions, do not exhibit much variation in skin tone.} {Finally, our face model tends to yield smooth 3D shape reconstructions, sometimes attributing fine shape details such as wrinkles on the face to the albedo instead of the shape. We believe that this is related to our shape generator following a convolutional architecture, which promotes local consistency between neighboring vertices of the face mesh.}

\section{Conclusion}
\label{sec:conclusion}
In this work we introduce MOST-GAN, a novel framework for manipulating face images in a 3D controllable and fully disentangled way. We achieve this by combining the physically-grounded modeling of 3DMMs with the expressive power of style-based GANs. We employ an encoder-decoder style architecture built on a 3DMM template, where we represent 3D shape, albedo, pose, and lighting independently by design. By coupling our 3D face model with a state-of-the-art 2D hair model, we develop a full portrait image manipulation pipeline. Unlike state-of-the-art methods, which require costly optimizations before manipulating real images, our method enables efficient image manipulation at inference time. Our results demonstrate the ability of our method to photorealistically manipulate 3D shape, albedo, pose, and lighting of face images, facilitating larger variations compared to state-of-the-art methods, and achieving better disentanglement in face image manipulation tasks.

\section*{Acknowledgements}
S. Medin worked on this project when he was an intern at MERL. B. Egger was supported by a PostDoc Mobility Grant, Swiss National Science Foundation P400P2\_191110.

\small
\bibliography{egbib}

\appendix
\twocolumn[\LARGE \centering \textbf{Supplementary Material} \vspace{5mm}]

\section{Architecture Details}
\textbf{Encoders.} For encoders $\{ \Eshape, \Ealbedo, \Elight, \Epose \}$, we employ the ResNet-18 architecture~\cite{he2016deep} (starting from the ImageNet~\cite{deng2009imagenet} pretrained weights) where we change the final layer to reflect the dimensionality of each latent representation: \smash{$\shapecode \in \mathbb{R}^{150}$}, \smash{$\albedocode \in \mathbb{R}^{200}$}, \smash{$\lightparam \in \mathbb{R}^{27}$}, and \smash{$\poseparam \in \mathbb{R}^{6}$}. 

\textbf{Generators.} For the albedo generator $\Galbedo$, we employ the original StyleGAN2~\cite{Karras2019stylegan2} architecture up to the $256 \times 256$ layer. (We omit the final layers with higher resolutions.) For the shape generator $\Gshape$, we use the architecture shown in Table~\ref{tab:shape-generator}. The output of this network is a UV-representation of shape from which we sample points corresponding to the UV-coordinates of each vertex in the FLAME~\cite{FLAME:SiggraphAsia2017} topology. Then, we add these as an offset to the FLAME mean shape to obtain a 3D shape in Euclidean space.

\textbf{Refiner.} For the refiner, we employ a U-Net~\cite{ronneberger2015u} with $5$ convolutional layers followed by $5$ transpose convolutional layers with skip connections. We provide the U-Net architecture details in Table~\ref{tab:refiner}.

\section{Training Details}

Throughout this section, we let $L_G$ denote the \emph{generator loss}, which is minimized to fool a discriminator that is trained adversarially to distinguish between generated face images and real face images~\cite{goodfellow2014generative}.

\textbf{Pretraining on Synthetic Data.} Our disentangled face model allows for pretraining each of the physical attributes separately. We carry out pretraining in three independent phases: albedo-only, lighting-only, and shape \& pose jointly. Using the notation we have introduced before, we minimize the following loss functions for the three phases:
\begin{align*}
    &\text{shape \& pose: } \hspace{-10pt} &&0.1 L_\mathrm{image}^\mathrm{syn} + 1000 L_\mathrm{shape}^\mathrm{syn} \\
    & && \hspace{60pt} + 100 L_\mathrm{pose}^\mathrm{syn} + 1.0 \Vert \boldsymbol{\alpha} \Vert_2^2 \\
    &\text{albedo-only: }  \hspace{-10pt} &&L_{G} + 10 L_\mathrm{image}^\mathrm{syn} + 100 L_\mathrm{albedo}^\mathrm{syn} + 1.0 \Vert \boldsymbol{\beta} \Vert_2^2 \\
    &\text{lighting-only: } \hspace{-10pt} &&10 L_\mathrm{image}^\mathrm{syn} + 100 L_\mathrm{light}^\mathrm{syn} 
\end{align*}
For each of these phases, we set the batch size to $16$ and use the Adam optimizer~\cite{kingma2014adam}. $\Eshape$, $\Ealbedo$, $\Elight$, $\Epose$, and $\Gshape$ are all trained with a learning rate of $0.0001$. During the albedo-only phase, we alternate optimization steps between training the albedo generator $\Galbedo$ and the image discriminator (both with learning rate $0.002$).

\textbf{Training on Real Data.} In the training stage, we split the FFHQ dataset~\cite{karras2019style} into train and test sets with $90\%-10\%$ split, using the first $63,\!000$ face images for training and the last $7,\!000$ images for testing. During training, we optimize over all networks in our face model ($\Eshape$, $\Ealbedo$, $\Elight$, $\Epose$, $\Gshape$, $\Galbedo$) jointly in an end-to-end fashion, and we alternate optimization steps between the face model and the image discriminator. For the first $50,\!000$ iterations, we minimize loss function~\eqref{eq:train_loss} for all blocks in the face model, using a batch size of $16$ and the Adam optimizer with learning rate $0.00001$. We use a batch size of $16$ and a learning rate of $0.00001$ for the discriminator as well.
\begin{multline}
    L_{G} + 1000 L_\mathrm{image}^\mathrm{real} + 10 L_\mathrm{identity}^\mathrm{real} + 100 L_\mathrm{landmark}^\mathrm{real}  \\
    + 1.0 L_\mathrm{albedo}^\mathrm{real}
    + 10^{-5} L_\mathrm{lighting}^\mathrm{real} + 1.0 \Vert \boldsymbol{\alpha} \Vert_2^2 + 1.0 \Vert \boldsymbol{\beta} \Vert_2^2
    \label{eq:train_loss}
\end{multline}
After training the network for $50,\!000$ iterations, we fine tune $\Ealbedo$, $\Elight$, $\Galbedo$ for another $50,\!000$ iterations by freezing the weights of $\Eshape, \Epose, \Gshape$ and discarding the landmark loss to further improve the reconstruction quality.

\textbf{Refinement.} Denoting the combined face-and-hair reconstruction as $\mathbf{\hat{x}_c} := \reconstface \odot \facemask + \reconsthair \odot \hairmask$, the refined image as $\finalimage := \Refiner(\mathbf{\hat{x}_c})$, and the original face and hair as $\mathbf{x'} := \inputimage \odot (\facemask + \hairmask)$, we employ the following loss function for the refiner:
\begin{multline}
    L_{G} + 8.0 \Vert f_\mathrm{VGG}(\mathbf{x'}) - f_\mathrm{VGG}(\finalimage) \Vert_2^2  \\ 
    + 10 \big(1 - \cos (f_\mathrm{id}(\mathbf{x'}), f_\mathrm{id}(\finalimage) ) \big)
\end{multline}
where the second term stands for the VGG-16 perceptual loss~\cite{simonyan2014very, johnson2016perceptual}. With a batch size of $16$, we alternately train the refiner and the discriminator for $500,\!000$ iterations, with learning rate $0.0001$ (using the Adam optimizer). To prevent overfitting, we randomly translate $\mathbf{x'}$ and $\mathbf{\hat{x}_c}$ together (with horizontal and vertical translations uniformly sampled from the range $[-15, 15]$ pixels).

\begin{table*}[t] \small
    \caption{\small Architecture of the shape generator $\Gshape$. The output of the network is a UV-representation of 3D shape, where the three channels of the $256 \times 256$ output represent 3D offsets (in $x$, $y$, and $z$) from the FLAME mean head shape~\cite{FLAME:SiggraphAsia2017}.}
\begin{tabularx}{\textwidth}{*{4}{>{\centering\arraybackslash}X}}
\cmidrule(r){1-4}
\textbf{layer type} & \textbf{kernel size / stride}  & \textbf{output shape} & \textbf{activation}  \\ \cmidrule(r){1-4}
linear & -- & $1024 \times 1$ & none \\
reshape &  -- & $16 \times 8 \times 8$ & -- \\ 
conv2d &  $4 \times 4 \ \ / \ \ 1$  & $32 \times 8 \times 8$ & tanh \\
upsample & -- & $32 \times 16 \times 16$ & -- \\ 
conv2d & $4 \times 4 \ \ / \ \  1$ &  $64 \times 16 \times 16$ & tanh \\ 
upsample & -- &  $64 \times 32 \times 32$&  --\\
conv2d & $4 \times 4 \ \  / \ \  1$ &  $64 \times 32 \times 32$& tanh \\ 
upsample & -- &  $64 \times 64 \times 64$ &  --\\ 
conv2d & $4 \times 4 \ \  / \ \  1$ & $64 \times 64 \times 64$& tanh \\ 
upsample & -- & $64 \times 128 \times 128$ &  --\\ 
conv2d & $4 \times 4 \ \  / \ \  1$ & $64 \times 128 \times 128$ & tanh \\ 
upsample & -- & $64 \times 256 \times 256$ &  --\\ 
conv2d & $4 \times 4 \ \  / \ \  1$ & $3 \times 256 \times 256$ & tanh  \\ \bottomrule
\end{tabularx}
\label{tab:shape-generator}
\end{table*}

\begin{table*}[t] \small
    \caption{\small Architecture of the refiner $\Refiner$. We employ a U-Net~\cite{ronneberger2015u} with skip connections between the encoder and decoder parts of the network. In all layers of the encoder, we use the LeakyReLU activation function with a negative slope of $0.2$. All layers of the decoder use the ReLU activation function.}
\begin{tabularx}{\textwidth}{*{4}{>{\centering\arraybackslash}X}}
\cmidrule(r){1-4}
\textbf{layer type} & \textbf{kernel size / stride}  & \textbf{output shape} & \textbf{activation}  \\ \cmidrule(r){1-4}
conv2d &  $4 \times 4 \ \ / \ \ 2$  & $64 \times 128 \times 128$ & LeakyReLU \\
conv2d & $4 \times 4 \ \  / \ \ 2$ & $128 \times 64 \times 64$ & LeakyReLU \\
conv2d & $4 \times 4 \ \  / \ \ 2$  & $256 \times 32 \times 32$ & LeakyReLU \\
conv2d & $4 \times 4 \ \ / \ \ 2$  & $512 \times 16 \times 16$ & LeakyReLU \\
conv2d & $4 \times 4 \ \ /\ \  2$  & $512 \times 8 \times 8$ & LeakyReLU \\
conv2d\_transpose & $4 \times 4 \ \ / \ \ 2$ & $512 \times 16 \times 16$ & ReLU \\ 
conv2d\_transpose & $4 \times 4 \ \ / \ \ 2$ & $256 \times 32 \times 32$ & ReLU \\ 
conv2d\_transpose & $4 \times 4 \ \ / \ \ 2$ & $128 \times 64 \times 64$ & ReLU \\ 
conv2d\_transpose & $4 \times 4 \ \ / \ \ 2$ & $64 \times 128 \times 128$ & ReLU \\ 
conv2d\_transpose & $4 \times 4 \ \ / \ \ 2$ & $3 \times 256 \times 256$ & ReLU \\ \bottomrule
\end{tabularx}
\label{tab:refiner}
\end{table*}

\section{Hair Manipulation Algorithm Details}

\textbf{Warp field calculation.} In each iteration of our hair manipulation algorithm, we first identify the visible triangles of the given face mesh with its reference pose, and compute the center of each triangle by taking the average of its vertices. Then, we project these triangle centers onto the image plane under both the reference and the target pose. Using the correspondences between the two projections, we construct a 2D warp by calculating how much each of the projected triangle centers moves in pixel space as a result of the pose change. To complete the warp field, we use the technique described below. 

For the vertical component of the warp field, we simply copy the vertical warp component from the nearest neighbor that was assigned a warp. For the horizontal component, we use a heuristic to assign a fixed horizontal warp to every pixel on the left edge of the image and a different fixed horizontal warp to every pixel on the right edge; then the horizontal component of the warp field for the entire image is simply interpolated from the assigned warps. In particular, when we rotate the faces clockwise (counter-clockwise) around the vertical axis, we extend a ray from the center of the 3D face to the left (right) perpendicular to the face's plane of symmetry and identify the 3D point on the ray whose projection lies on the leftmost (rightmost) edge of the image. Next, we calculate by how much this point's projection into the image plane moves when the head rotates, and we multiply this number by $3$ to obtain the horizontal warp that we assign to the leftmost (rightmost) column of the warp field. For the rightmost (leftmost) column, we heuristically choose a displacement of $10$ pixels to the right (left). Finally, we interpolate between the assigned pixels using linear interpolation to obtain the horizontal warp of every image pixel.

\textbf{Regularization of the hair mask.} After we obtain a complete warp field, we apply it to the reference hair mask and orientation. The orientation is regularized as part of the MichiGAN~\cite{tan2020michigan} pipeline, whereas we regularize the mask by projecting it onto a PCA basis that we calculate from a dataset of hair masks. In particular, we construct our hair mask dataset by randomly selecting $10,\!000 $ samples from our FFHQ training set and combining it with $10,\!000$ masks that we obtain from the USC Hair Salon database~\cite{hu2015single}. For the latter, we attach 3D hair models from the USC Hair Salon database to the FLAME mean head model, which we rotate around the vertical axis by an angle uniformly sampled from $[-90^\circ, 90^\circ]$. The hair masks are obtained by rendering these 3D shapes. Finally, after downsampling all masks to $64\times64$ resolution, we construct a PCA basis with $50$ principal components, onto which we project the hair masks at each iteration of our reposing algorithm. In this work, starting from the pose of the original face image, we invoke this algorithm sequentially by imposing a pose change of $5^\circ$ in each iteration, going all the way to the full profile ($\pm 90^\circ$) poses.

\begin{figure*}[t]
    \centering
    \includegraphics[trim=0mm 0mm 0mm 0mm,clip,width=0.75\textwidth]{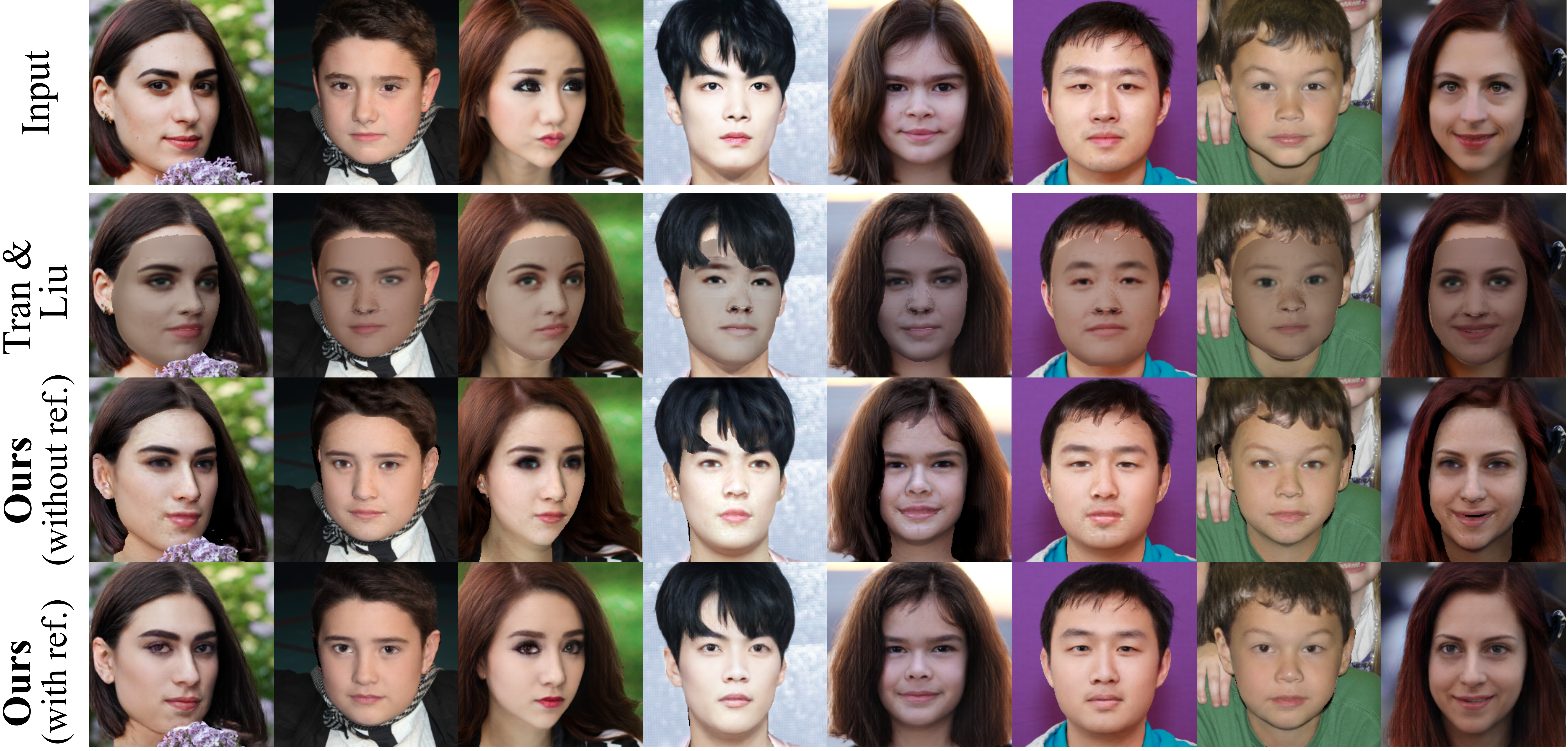}
\caption{\small Ablation study and comparison with~\citet{tran2019learning}. On images from the test set (never seen during training), our method is able to reconstruct faces more accurately and photorealistically than~\citet{tran2019learning}. These results also demonstrate that our full model (with refinement) shows a notable improvement in image quality vs.\ our model without the refiner.
} 
\label{fig:comparison-tran-liu}
\end{figure*}
\begin{table*}[t] \small
    \caption{\small Average face recognition (FR) scores, SSIM and PSNR between the original and reconstructed face images from our dataset. We observe a notable improvement due to the refinement.}
\begin{tabularx}{\textwidth}{l*{3}{>{\centering\arraybackslash}X}}
\cmidrule(r){1-4}
& FR score $\uparrow$ & SSIM $\uparrow$ & PSNR $\uparrow$   \\ \cmidrule(r){1-4}
{Ours (without refinement)} & ${0.66 \pm 0.10}$ & ${0.72 \pm 0.09}$ & ${22.24 \pm 2.63}$ \\
\textbf{Ours (with refinement)} & $\mathbf{0.68 \pm 0.10}$ & $\mathbf{0.74 \pm 0.08}$ & $\mathbf{23.20 \pm 2.47}$  \\ \bottomrule
\end{tabularx}
\label{tab:ablation}
\end{table*}
\begin{table*}[t!] \small
    \caption{\small Average face recognition (FR) scores, SSIM and PSNR between the original and reconstructed face images for our method (both with and without refinement) and~\citet{tran2019learning}. To obtain the results in this table, we masked out the hair, background, clothing, and teeth for fair comparison with~\citet{tran2019learning}. Our method achieves better scores in all three metrics.}
\begin{tabularx}{\textwidth}{l*{3}{>{\centering\arraybackslash}X}}
\cmidrule(r){1-4}
& FR score $\uparrow$ & SSIM $\uparrow$ & PSNR $\uparrow$   \\ \cmidrule(r){1-4}
\citet{tran2019learning} & $0.51 \pm 0.12$ & $0.87 \pm 0.03$ & $20.96 \pm 1.57$ \\ 
{Ours (without refinement)} & ${0.69 \pm 0.10}$ & ${0.87 \pm 0.04}$ & ${25.08 \pm 2.92}$  \\
\textbf{Ours (with refinement)} & $\mathbf{0.71 \pm 0.10}$ & $\mathbf{0.88 \pm 0.04}$ & $\mathbf{26.17 \pm 2.71}$  \\
\bottomrule
\end{tabularx} 
\label{tab:comparison-tran-liu}
\end{table*}

\section{Ablation Study}
{\textbf{Training and loss functions.} In our experiments, we observed that pretraining on synthetic data training is crucial for our method to work, since we observed stability issues when we started out by training on real data. For real data training, our experiments suggested that all loss functions except for equations (11)--(13) in the paper (all loss functions except for the albedo regularization, lighting regularization, and regularization of shape and albedo codes) are crucial for our method to achieve reasonable face reconstructions. When we omitted albedo and lighting regularizations, we observed that our method converges to a state in which the albedo reconstructions are washed out and the appearance of the face is mostly attributed to the lighting, which suggests that albedo and lighting regularizations are important to achieve better albedo and lighting disentanglement.}

{\textbf{Refinement.}} In this section, we analyze the impact of the refiner on our reconstructions by providing qualitative and quantitative comparison of our results with vs.\ without the refiner. In addition, we compare our reconstructions with a state-of-the-art nonlinear 3D morphable model proposed by Tran and Liu~\cite{tran2019learning}. We illustrate our qualitative comparisons in Figure~\ref{fig:comparison-tran-liu}. It is clear that our reconstructions are much more accurate and photorealistic than those of Tran and Liu. We also observe a notable improvement in photorealism using our complete model (with refiner) vs. using our model without the refiner. To quantify our observations, we calculate a face recognition (FR) score as the average cosine similarity between the feature vectors extracted from the ArcFace face recognition network~\cite{deng2018arcface} for the original and reconstructed images (from our test dataset). We also compute the structural similarity index measure (SSIM) and peak signal-to-noise ratio (PSNR)~\cite{hore2010image} between the original and reconstructed images. We quanitatively compare our model with vs. without the refiner in Table~\ref{tab:ablation}. In Table~\ref{tab:comparison-tran-liu}, we perform quantitative comparisons with Tran and Liu. To obtain the results in Table~\ref{tab:comparison-tran-liu}, we masked out the hair, background, clothing, and teeth for fair comparison with Tran and Liu~\cite{tran2019learning}.

\section{Additional Experiments and Results}

In this section, we provide additional experiments and qualitative results.

\begin{figure*}[t]
    \centering
    \includegraphics[trim=0mm 0mm 0mm 0mm,clip,width=0.75\textwidth]{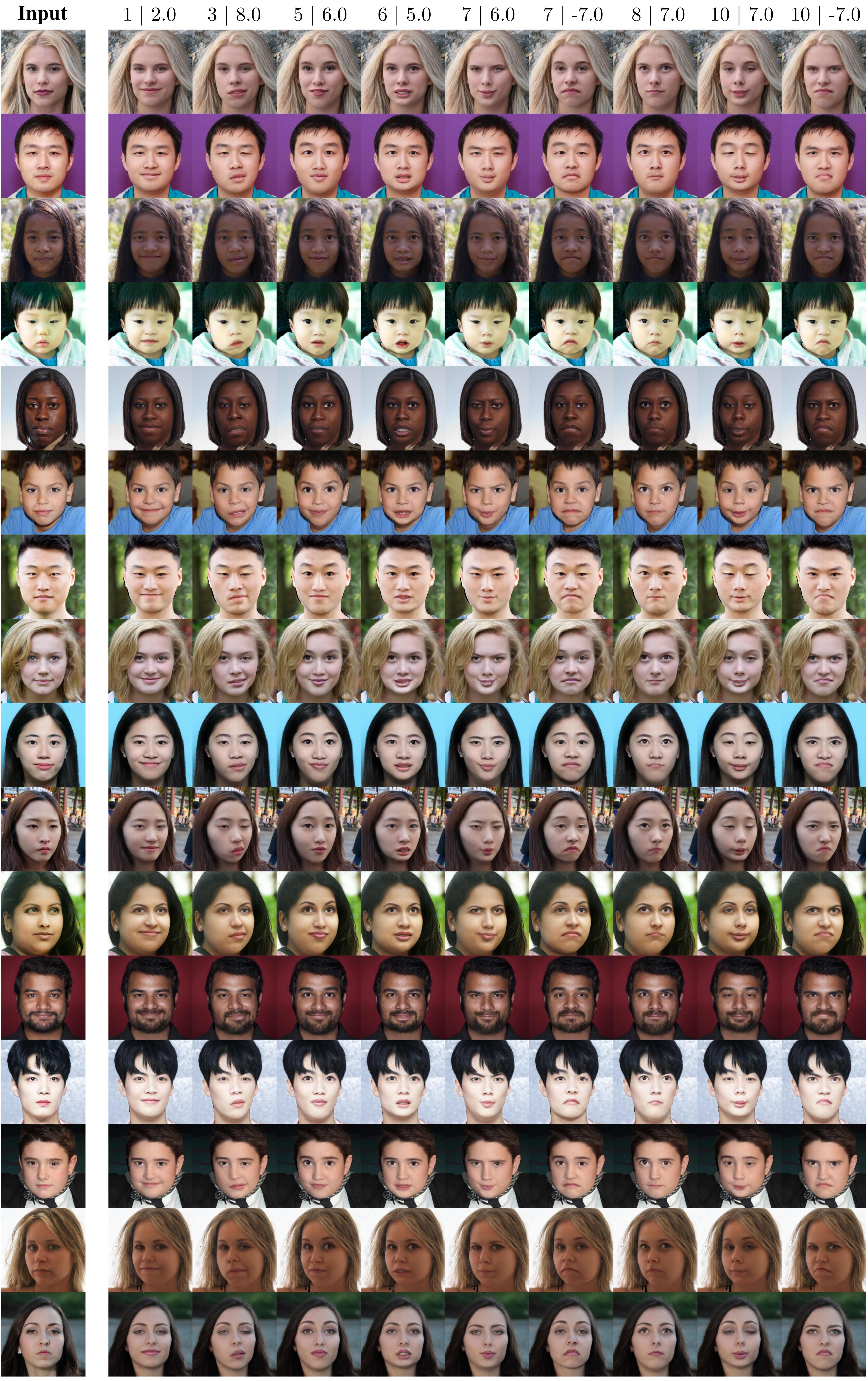}
    \caption{\small \textbf{Expression manipulation results.} We illustrate several expression changes with varying intensities. The two numbers above each column indicate which FLAME expression eigenvector is used and by how many standard deviations it is scaled.} 
    \label{fig:expr-manip}
\end{figure*}

\textbf{Expression manipulation.} In Fig.~\ref{fig:expr-manip}, we present additional expression manipulation results, generated using the same method described in the main paper (see Fig.~\ref{fig:expr-light-manip} in the main paper).

\begin{figure*}[t]
    \centering
    \includegraphics[trim=0mm 0mm 0mm 0mm,clip,width=0.75\textwidth]{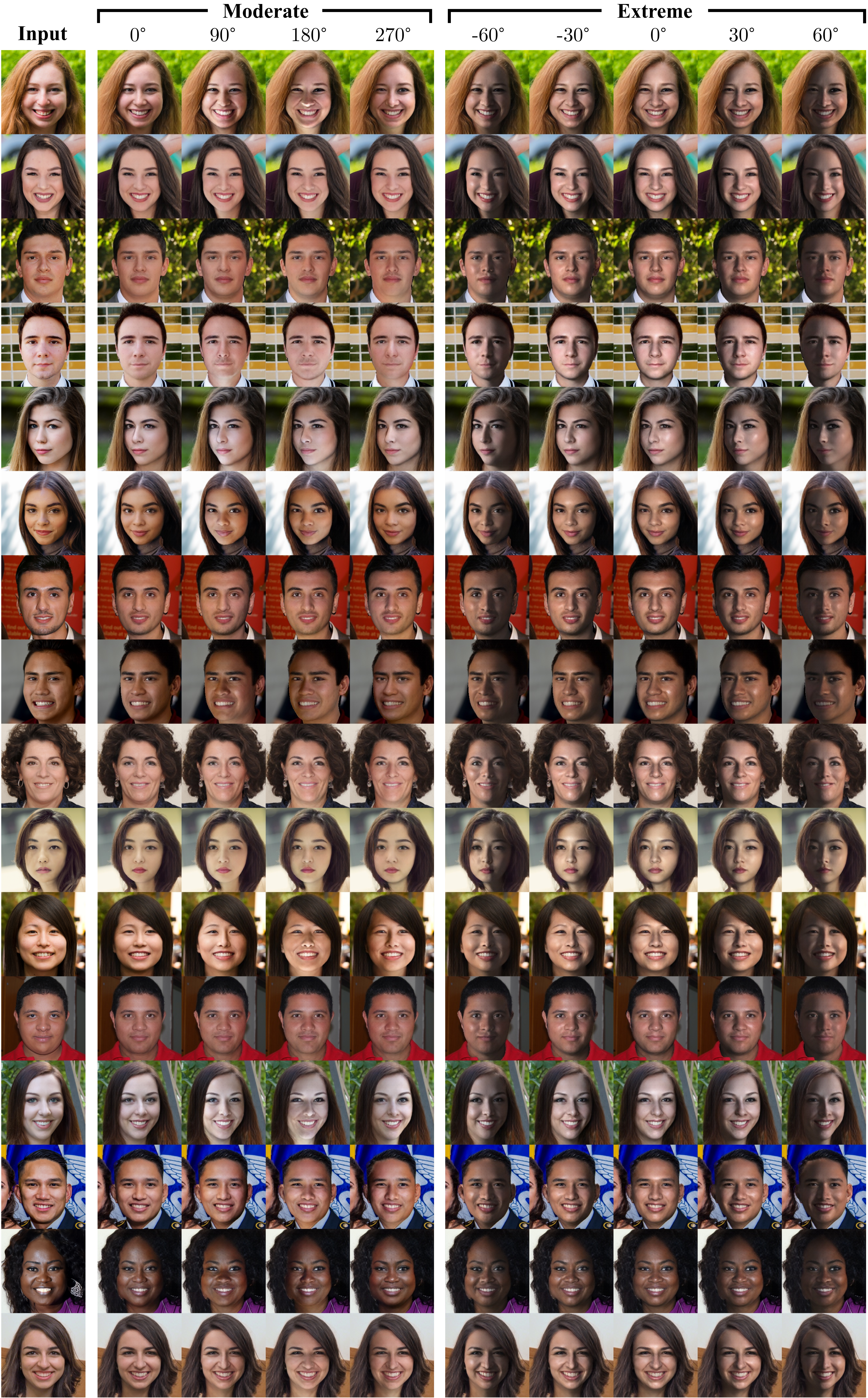}
    \caption{\small \textbf{Lighting manipulation results.} For moderate variation, we rotate the reconstructed lighting around the camera axis by the angle listed above each column. For extreme lighting, we render the reconstructed 3D model using a point light source and Phong shading model.} 
    \label{fig:light-manip}
\end{figure*}

\textbf{Lighting manipulation.} In Fig.~\ref{fig:light-manip}, we present additional lighting manipulation results, generated using the same method described in the main paper (see Fig.~\ref{fig:expr-light-manip} in the main paper).

\begin{figure*}[t]
    \centering
    \includegraphics[trim=0mm 0mm 0mm 0mm,clip,width=0.75\textwidth]{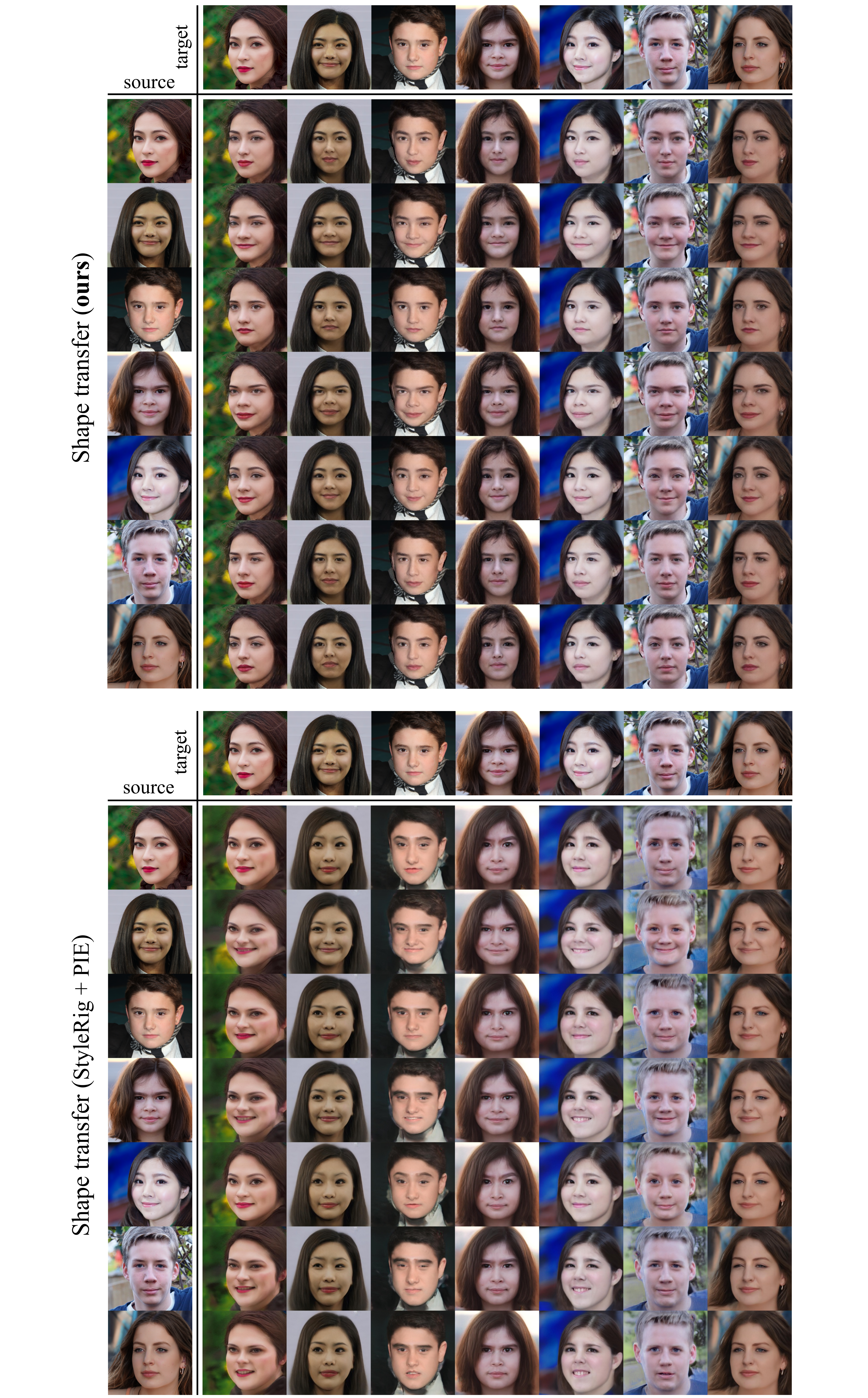}
    \vspace{-0.2cm}
    \caption{\small \textbf{Shape transfer results and comparison}. We transfer the 3D shape of each source image to each target image while keeping everything else unchanged. Compared to the previous state of the art (StyleRig~\cite{tewari2020stylerig} + PIE~\cite{tewari2020pie}), our results (\emph{top}) demonstrate more accurate shape transfer and much better disentanglement between shape and other face attributes (e.g., albedo, pose, and hair).
    } 
    \vspace{-3mm}
    \label{fig:shape-transfer-supp}
\end{figure*}

\textbf{Shape transfer.} In Fig.~\ref{fig:shape-transfer-supp}, we present additional shape transfer results, generated using the same method described in the main paper (see Fig.~\ref{fig:shape-transfer} in the main paper), where we also compare with results obtained by a combination of StyleRig~\cite{tewari2020stylerig} and PIE~\cite{tewari2020pie}. In this combination of previous methods, the image embedding is carried out by the optimization algorithm proposed in PIE, and the shape transfer is performed using StyleRig.

\begin{figure*}[t]
    \centering
    \includegraphics[trim=0mm 0mm 0mm 0mm,clip,width=0.80\textwidth]{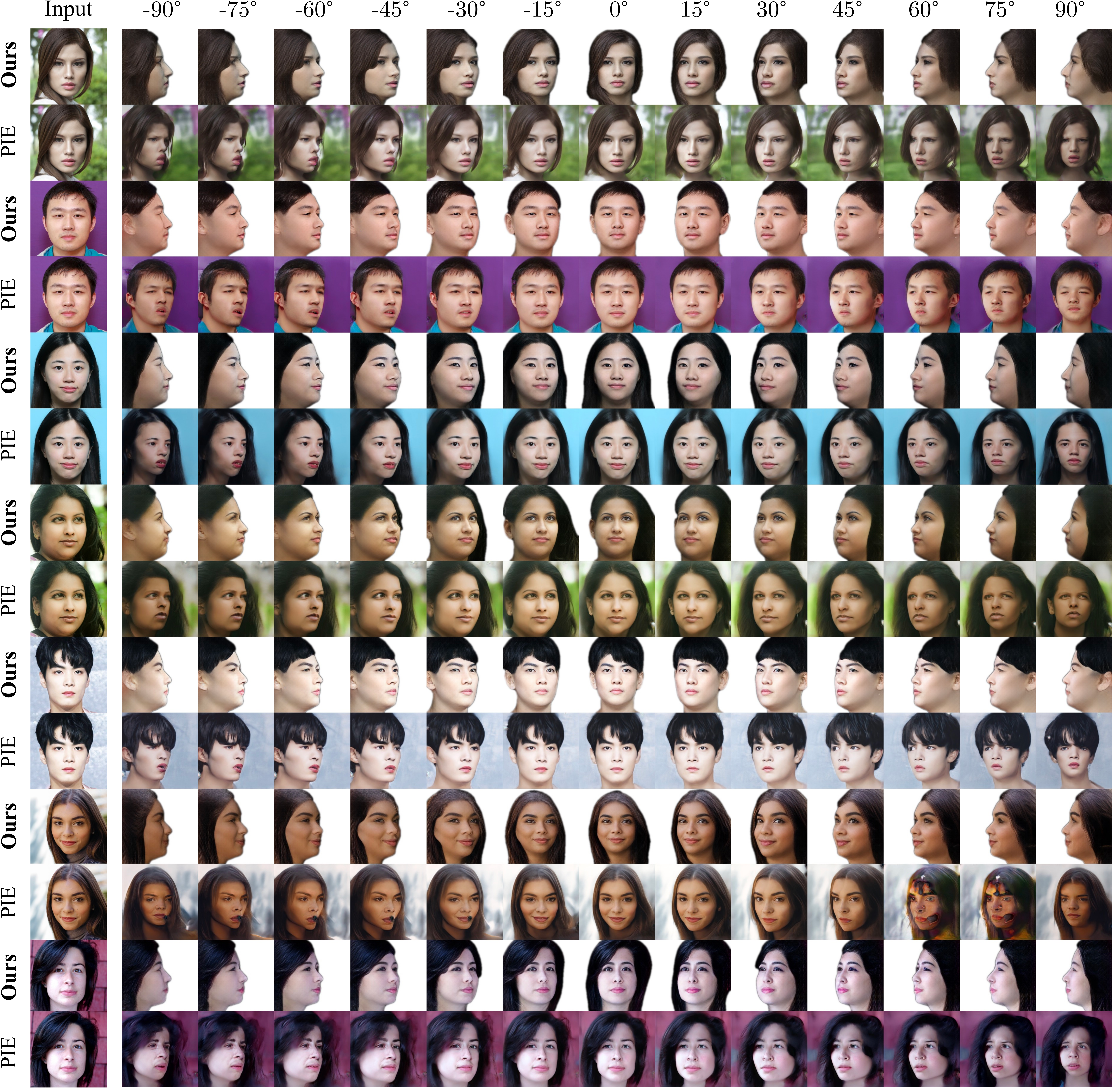}
    \vspace{-0.2cm}
    \caption{\small \textbf{Pose manipulation results and comparison with PIE}~\cite{tewari2020pie}. To edit the pose of a given portrait image, we rotate the reconstructed faces in 3D and warp the hair in 2D. Our method is able to rotate portrait images all the way to profile pose while keeping the identity, expression, and illumination conditions unchanged. 
    } 
    \vspace{-3mm}
    \label{fig:pose-manip-supp}
\end{figure*}

\textbf{Pose manipulation.} In Fig.~\ref{fig:pose-manip-supp}, we present additional pose manipulation results, generated using the same method described in the main paper (see Fig.~6 in the main paper).

\textbf{Joint transfer of physical attributes.} 
Our face model's full disentanglement is demonstrated by its ability to transfer all physical attributes either individually or jointly. In Fig.~\ref{fig:shape-texture-transfer}, we present results of joint albedo and lighting transfer, as well as results of joint transfer of albedo, lighting, and shape.

\begin{figure*}[t]
    \centering
    \includegraphics[trim=0mm 0mm 0mm 0mm,clip,width=0.80\textwidth]{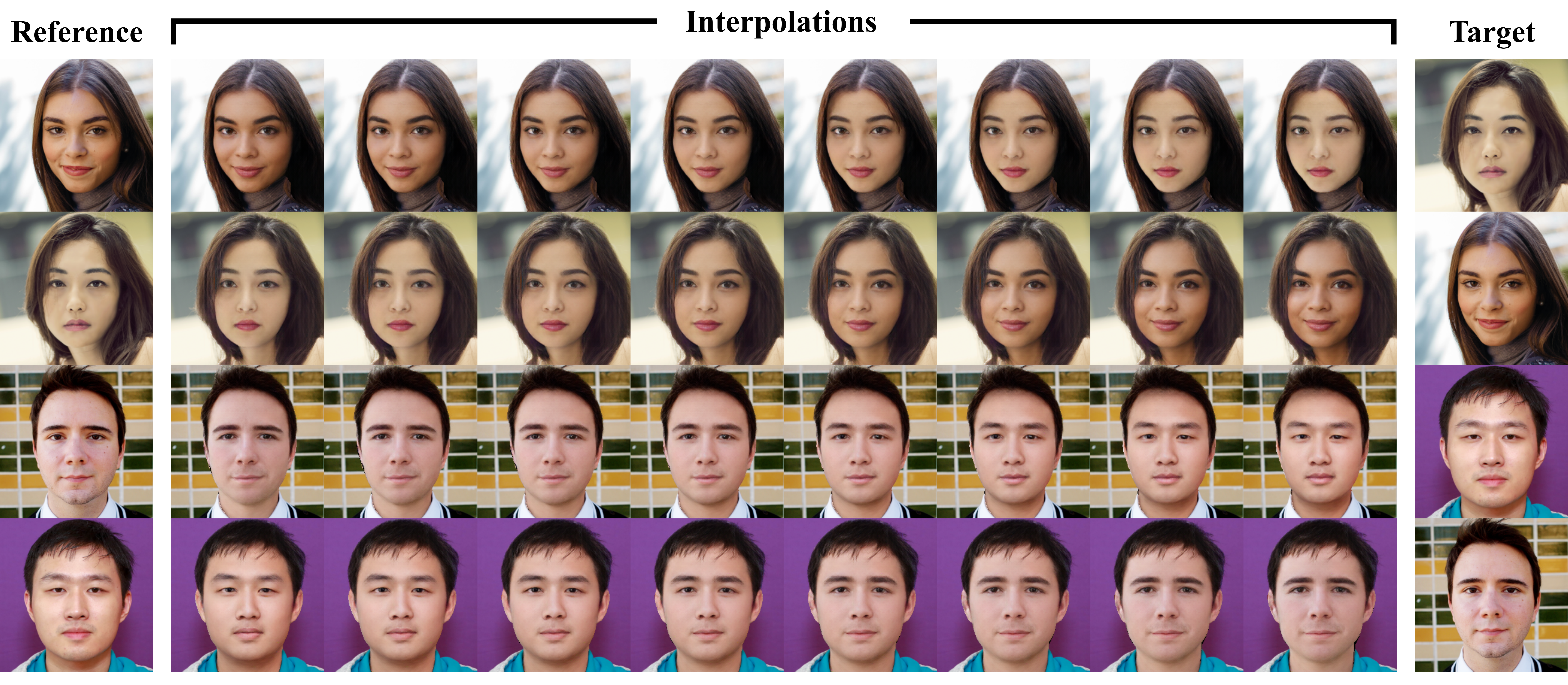}
    \vspace{-0.2cm}
\caption{\small \textbf{Interpolation results.} Our method also allows for interpolation in the latent spaces. In each row, given a reference and a target image, we interpolate between their shape codes, albedo codes, and lighting parameters. For each interpolated image, the background and the latent code for hair are copied from the reference image.} 
    \vspace{-3mm}
    \label{fig:interpolation}
\end{figure*}

\begin{figure*}[t]
    \centering
    \includegraphics[trim=0mm 0mm 0mm 0mm,clip,width=0.85\textwidth]{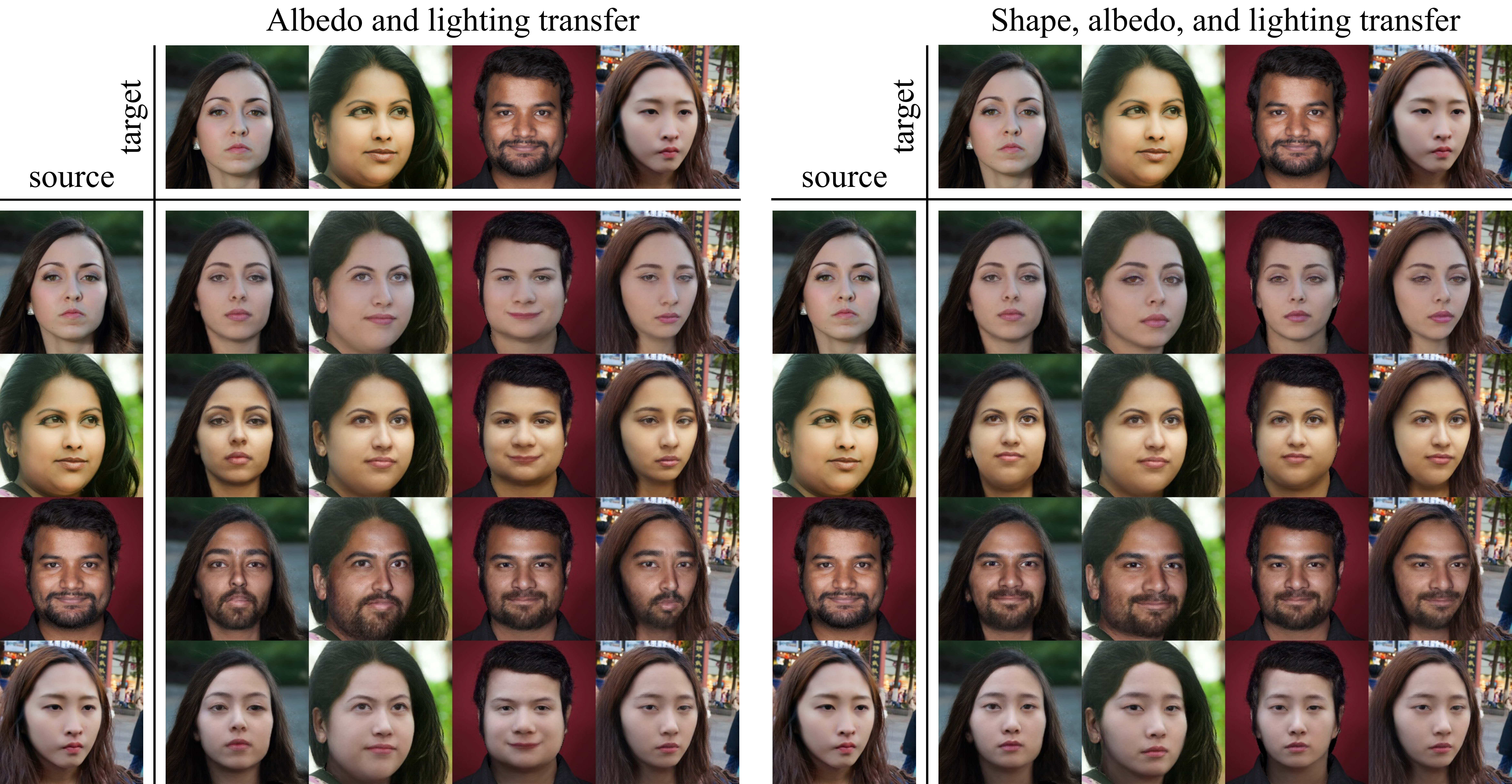}
    \vspace{-0.2cm}
\caption{\small \textbf{Joint transfer of physical attributes.} Our method is able to transfer physical attributes jointly as well as individually. In this figure, we jointly transfer the indicated physical attributes of each source image to each target image while keeping the other parameters of the target image unchanged. {\em Left:} Albedo and lighting transfer. {\em Right:} Shape, albedo, and lighting transfer.}

    \vspace{-3mm}
    \label{fig:shape-texture-transfer}
\end{figure*}

\textbf{Interpolation in the latent space.} Although we do not impose any smoothness constraints within the latent spaces, the learned shape and albedo latent spaces enable smooth interpolation between different latent codes. We present our interpolation results in Fig.~\ref{fig:interpolation}, where we simultaneously interpolate between the reconstructed shape code, albedo code, and lighting parameters of the reference and target images.

\begin{figure*}[t]
    \centering
    \includegraphics[trim=0mm 0mm 0mm 0mm,clip,width=0.85\textwidth]{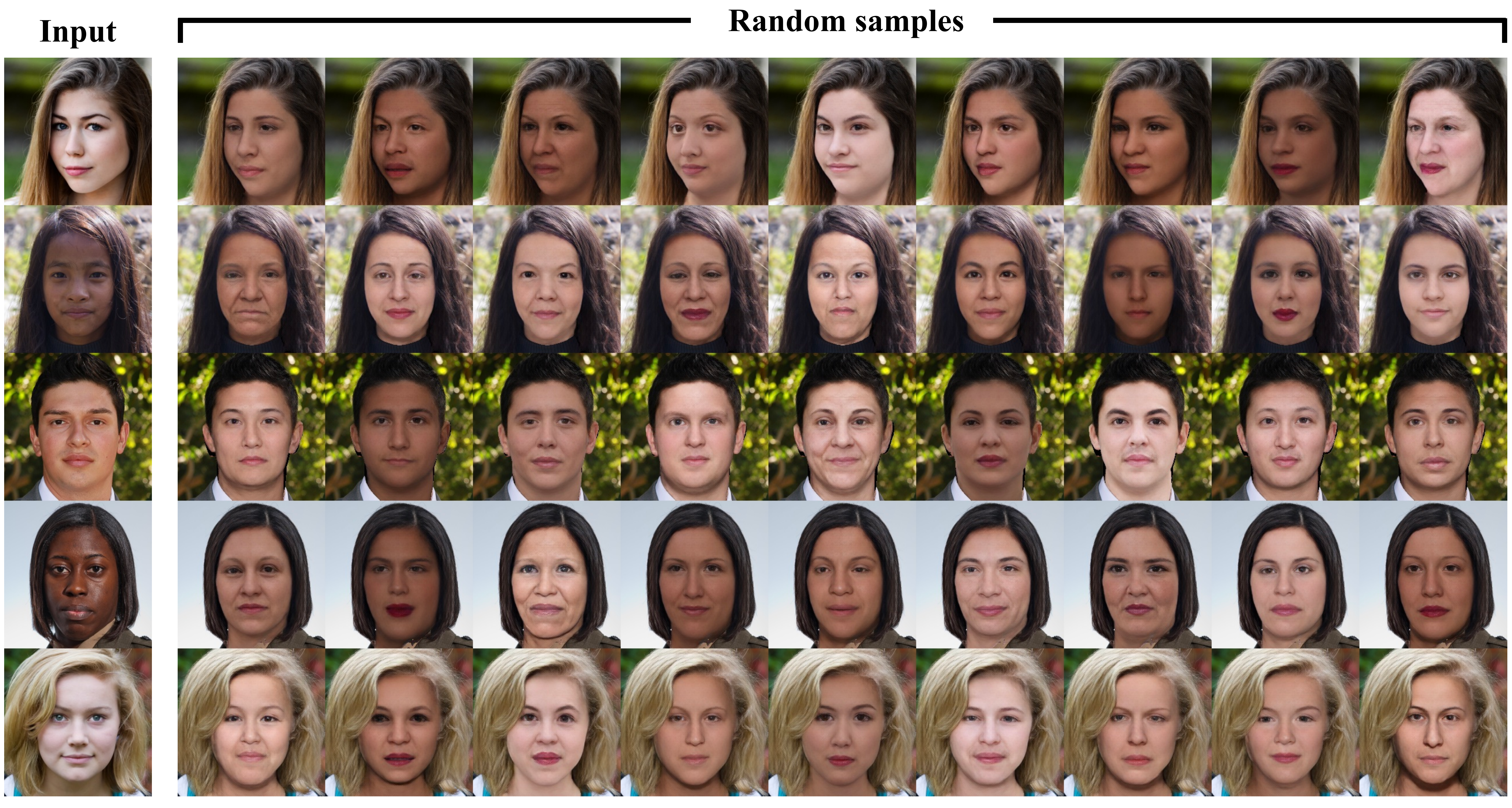}
    \vspace{-0.2cm}
    \caption{\small \textbf{Face anonymization results.} Our model can also be used to sample novel faces by regularizing the latent code distributions during training, which can be used for face anonymization.}
    \vspace{-3mm}
    \label{fig:random-samples}
\end{figure*}

\textbf{Face anonymization.} Our face model can also be turned into a generative model for random faces by regularizing the latent code distributions during training. At each iteration, we calculate sample means and variances of the latent codes for shape and albedo over minibatches and regularize these statistics to match those of the standard Gaussian distribution by using a KL divergence loss. After this regularized training, we can randomly sample codes from a standard multivariate Gaussian distribution to produce realistic shapes and albedos. To enable random sampling of lighting conditions that match the distribution of lighting conditions present in the training set, we train a variational autoencoder on the reconstructed lighting parameters from the training set. In Fig.~\ref{fig:random-samples}, we use random sampling of latent codes (i.e., random face generation) to anonymize face images from the test set. In each row of the figure, the input image is fed through our encoders to determine the pose and latent hair code. To anonymize the face, we randomly sample latent codes for shape, albedo, and lighting, while retaining the input image's background, pose, and hair code.

\textbf{Video of our results.} We also provide a video of our results where we demonstrate smooth manipulations in expression, lighting, albedo, shape, and their combinations.

\end{document}